\documentclass[11pt]{article}

\usepackage[final]{acl}

\usepackage{times}
\usepackage{latexsym}

\usepackage[T1]{fontenc}

\usepackage[utf8]{inputenc}

\usepackage{microtype}

\usepackage{inconsolata}

\usepackage{graphicx}
\usepackage{amsmath}
\usepackage{subcaption}
\usepackage{textcomp}

\usepackage{multirow}
\usepackage{graphicx}
\usepackage[normalem]{ulem}
\useunder{\uline}{\ul}{}

\usepackage{amssymb}
\usepackage{tcolorbox} 
\usepackage{enumitem} 
\usepackage{url}
\usepackage[none]{hyphenat}
\usepackage{cuted}
\usepackage{eso-pic}

\title{LLM-MC-Affect: LLM-Based Monte Carlo Modeling of Affective Trajectories and Latent Ambiguity for Interpersonal Dynamic Insight}

\author{
    \textbf{Yu-Zheng Lin\textsuperscript{1}\thanks{Corresponding Author}}, 
    \textbf{Bono Po-Jen Shih\textsuperscript{2}}, 
    \textbf{John Paul Martin Encinas\textsuperscript{1}}, \\
    \textbf{Elizabeth Victoria Abraham Achom\textsuperscript{1}}, 
    \textbf{Karan Himanshu Patel\textsuperscript{1}}, 
    \textbf{Jesus Horacio Pacheco\textsuperscript{3}}, \\
    \textbf{Sicong Shao\textsuperscript{4}}, 
    \textbf{Jyotikrishna Dass\textsuperscript{1}}, 
    \textbf{Soheil Salehi\textsuperscript{1}}, 
    \textbf{Pratik Satam\textsuperscript{1}}
\\
\\
 \textsuperscript{1}University of Arizona 
 \textsuperscript{2}Pennsylvania State University \\
 \textsuperscript{3}Universidad de Sonora
 \textsuperscript{4}University of North Dakota
\\
 \small{
   \textbf{*Correspondence:} \href{mailto:email@domain}{yuzhenglin@arizona.edu}
 }
}

\begin{document}
\AddToShipoutPictureBG*{%
  \AtPageUpperLeft{%
    \raisebox{-0.55in}{%
      \makebox[\paperwidth][c]{%
        \small Accepted to the 64th Annual Meeting of the Association for Computational Linguistics (ACL 2026)
      }%
    }%
  }%
}

\maketitle

\begin{abstract}
\begin{figure*}[h!]
\centering
\includegraphics[width=.82\textwidth]{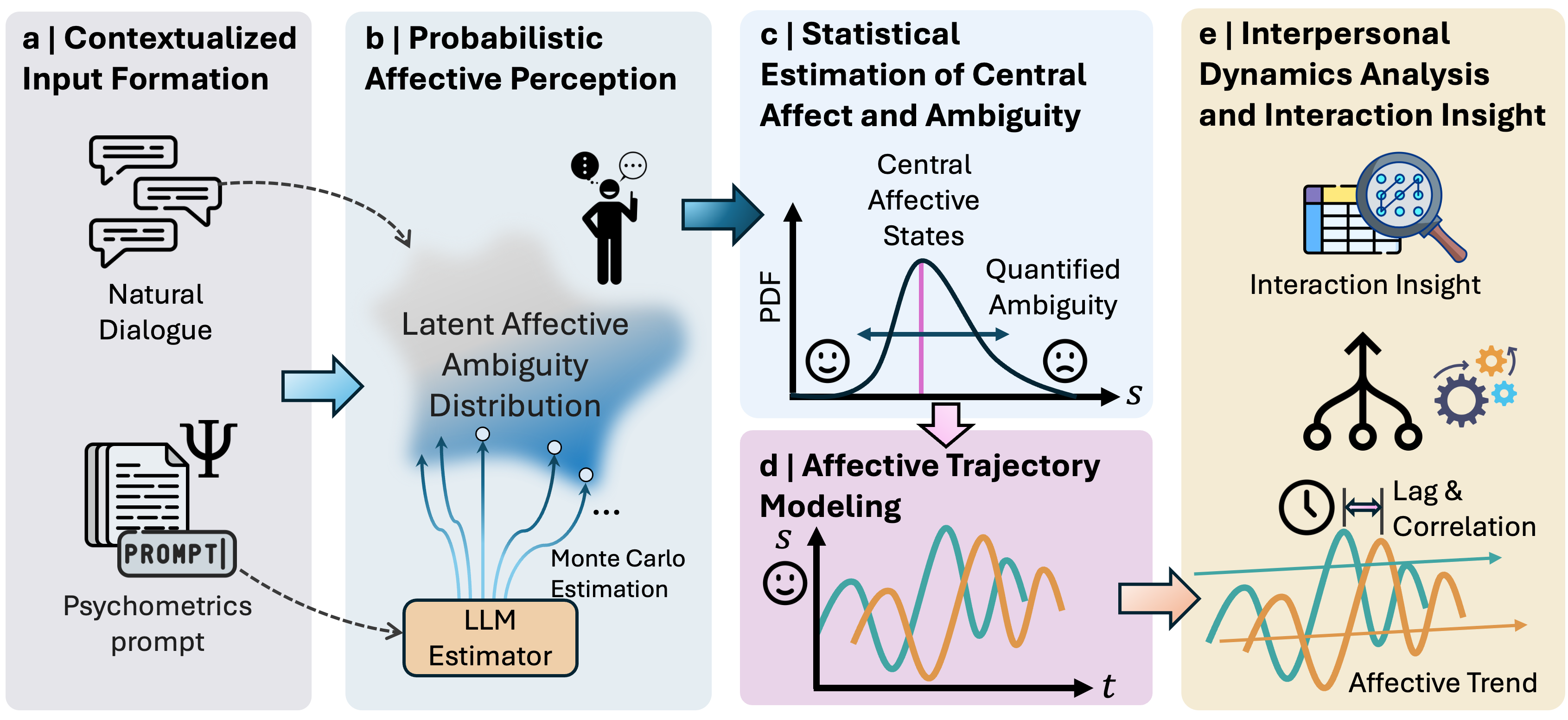}
\caption{Overview of the LLM-MC-Affect framework for probabilistic affect modeling and dyadic interaction analysis. (a) Contextualized input formation integrates natural dialogue with a psychometric prompt to guide affect estimation. (b) Probabilistic affective perception leverages stochastic LLM decoding with Monte Carlo sampling to estimate a latent distribution over affective states. (c) Statistical estimation derives the central affective tendency and quantifies ambiguity from the inferred distribution. (d) Affective trajectory modeling tracks the temporal evolution of affective states, capturing both trends and uncertainty over time. (e) Interpersonal dynamics analysis jointly interprets lag, correlation, and trajectory trends to extract interpretable dyadic interaction patterns, linking low-level probabilistic affect measurements to higher-level interaction insights.}
\label{fig:Architecture}
\end{figure*}
Emotional coordination is a fundamental aspect of human interaction that shapes how relational meaning unfolds across interaction. While text-based affect inference has become increasingly feasible, prior approaches often treat sentiment as a deterministic point estimate for individual speakers, failing to capture the inherent subjectivity, latent ambiguity, and sequential coupling found in mutual exchanges. We introduce LLM-MC-Affect, a probabilistic framework that characterizes emotion not as a static label, but as a continuous latent probability distribution defined over an affective space. By leveraging stochastic LLM decoding and Monte Carlo (MC) estimation, the methodology approximates these distributions to derive high-fidelity sentiment trajectories that explicitly quantify both central affective tendencies and perceptual ambiguity. These trajectories enable a structured analysis of interpersonal coupling through sequential cross-correlation and slope-based indicators, identifying leading or lagging influences between interlocutors. To validate the interpretive capacity of this approach, we utilize teacher-student instructional dialogues as a representative case study, where our quantitative indicators successfully distill high-level interaction insights such as effective scaffolding. This work establishes a scalable and deployable pathway for understanding interpersonal dynamics, offering a generalizable solution that extends beyond education to broader social and behavioral research.
\end{abstract}

\section{Introduction}

Emotional coordination is a core property of human interaction that shapes how relational meaning is constructed in real time, particularly in instructional settings where rapport, engagement, and frustration can propagate across conversational turns. One related concept is affective synchrony \cite{wood2021forms}, which broadly refers to the alignment or co-evolution of affective states among interacting individuals over time. Empirical studies of affective synchrony have traditionally relied on biometric coupling signals \cite{qi2024dyadic,10.1093/scan/nsab103,bevilacqua2019brain}, such as autonomic responses or neural alignment, which provide high sequential fidelity but require specialized instrumentation and tightly controlled protocols that are difficult to scale to large, naturalistic learning environments. In contrast, conversational text is ubiquitous and inherently longitudinal, and prior studies have demonstrated that it encodes a recoverable affective structure that can be summarized as sentiment trajectories \cite{liu2024text}. However, many text-based sentiment analysis pipelines continue to represent affect as a deterministic point estimate at the level of individual utterances or speakers \cite{liu2023emotion, gao2022emotion}. Although effective for coarse-grained evaluation, this representation tends to compress subjective variability, offers limited visibility into latent affective ambiguity, and often treats affect as an individual attribute rather than an explicitly coupled dyadic process in interaction analysis \cite{tsakalidis-etal-2022-identifying, wemmer-etal-2024-emoprogress}.

This paper introduces LLM-MC-Affect, a probabilistic framework that treats utterance-level emotion not as a fixed label, but as a latent affective distribution over a continuous affective space. Recent advances in large language models have demonstrated strong zero-shot affective perception capabilities, enabling them to infer nuanced emotional states from natural language without task-specific fine-tuning \cite{lin2025personalized,zhang2025dialoguellm,liu2024emollms}. Rather than approximating this latent distribution through labels collected from multiple human raters, which can be costly and difficult to scale, we leverage the stochastic nature of large language models under non-zero decoding temperature \cite{renze2024effect,liu2023g,salinas2025tuning} and perform LLM-based Monte Carlo (MC) estimation: repeated, independently sampled affective evaluations are treated as draws from an implicit model-induced proxy for the latent distribution, enabling estimation of both (1) the central affective tendency and (2) dispersion as a quantitative indicator of latent affective ambiguity. This yields stable, uncertainty-aware sentiment trajectories that explicitly retain the variance information discarded by deterministic scoring. 

Building on these trajectories, LLM-MC-Affect characterizes interaction-level synchrony by modeling how two affective signals align across conversational turns. Specifically, we compute a normalized sequential cross-correlation function between teacher and student trajectories, estimate an influence lag that indicates leading or lagging influence, and combine these sequential alignment measures with slope-based trend indicators that summarize long-horizon affective development. By jointly interpreting lag, correlation sign, and trajectory slope coupling, the framework yields interpretable dyadic patterns (e.g., teacher-led positive scaffolding, synchronous decline, or dynamic compensation) derived from low-level probabilistic affect measurements, thereby linking fine-grained affective perception to higher-level interpersonal interpretation.

We validate the method on teacher-student instructional dialogues as a representative case study, because education provides a structured yet socially rich domain in which affective coupling is meaningful and can be qualitatively examined. Under a unified, prompt-defined psychometric rubric, the proposed approach supports zero-shot affect estimation without task-specific fine-tuning, while enabling controlled sensitivity analyses over decoding stochasticity and cross-model comparisons that reveal differences in affective resolution and ambiguity across LLMs. Taken together, the results demonstrate that affective synchrony and directional influence can be derived from linguistic affect alone, providing a deployable pathway for affect-aware instructional analytics and, more broadly, for scalable research on interpersonal dynamics without biometric sensing.

Our contributions are summarized as follows: (1) We propose LLM-MC-Affect, a probabilistic pipeline that uses stochastic LLM decoding with Monte Carlo estimation to model utterance-level affect as a latent distribution and quantify ambiguity via variance, rather than relying on deterministic sentiment labels. (2) We introduce an interaction analysis layer that extracts interpretable dyadic indicators, including sequential cross-correlation, estimated influence lag, and slope-based trajectory signals, yielding a structured typology of interpersonal affective dynamics. (3) We demonstrate feasibility in instructional dialogues, showing that uncertainty-aware trajectories remain robust under decoding stochasticity and can support interpretable affective synchrony insights in a scalable, text-only setting.

\section{Background}
\subsection{Statistical Formulation of Affective Ambiguity}
Affective modeling in natural language interaction must account for the inherent subjectivity and multifaceted ambiguity found in emotional expressions \cite{davani2022dealing, kuhn2023semantic}. Drawing upon the conceptual foundation established by \citet{11185123}, we characterize the emotion associated with a specific utterance $u_t$ not as a deterministic point estimate, but as a latent probability distribution $p(e|u_t)$ defined over a continuous affective space $\mathcal{E}$. This probabilistic paradigm acknowledges that the actual underlying continuous distribution $p(e|u_t)$ is a latent construct that remains inaccessible for direct observation. Consequently, conventional methodologies approximate this hidden continuous density by collecting a finite set of independent inferences, such as human ratings, where each individual's discrete score is a sample discretized from the continuous distribution $p(e|u_t)$. This statistical premise establishes that inter-rater variability and perceived ambiguity are not merely noise to be averaged out; instead, they represent essential emotional nuances that characterize the underlying affective state. 
\subsection{Emotion as a Temporally Coupled Interpersonal System}
Emotion in social interaction is increasingly understood as a temporally structured process that emerges through ongoing interpersonal coordination rather than as an isolated individual attribute. From this perspective, \citet{butler2011temporal} conceptualizes emotions as Temporal Interpersonal Emotion Systems (TIES), in which social partners function as a dynamically coupled, self-organizing system. In this framework, emotional subcomponents interact across individual boundaries over time, giving rise to integrated dyadic patterns. From an emotion dynamics perspective, cross-correlation functions provide an effective and well-established tool for capturing time-lagged relationships between affective processes, enabling the characterization of temporal coordination and relative lag in dyadic interactions after accounting for overall trends and level effects \cite{feldman2003infant, krone2018multivariate, main2016exploratory}.

\section{Methodology}
We introduce LLM-MC-Affect, a probabilistic framework for modeling affective trajectories from conversational text using LLMs. The method treats utterance-level emotion not as a deterministic label, but as a latent distribution estimated via Monte Carlo sampling under stochastic LLM decoding, enabling robust affective trajectory construction while explicitly capturing affective ambiguity. These trajectories form a continuous sequential representation that supports downstream interpersonal analysis, including sequential alignment, directional influence, and long-term affective progression. By aggregating low-level probabilistic sentiment estimates into interpretable interaction indicators, LLM-MC-Affect bridges fine-grained affect perception with higher-level explanatory insights. The overall processing pipeline is illustrated in Figure \ref{fig:Architecture}.

\subsection{LLM-Based Affective Trajectory Modeling}
\label{sec:llm_sentiment}
This subsection formulates affective trajectory modeling as a probabilistic inference problem over latent emotion distributions. Given a dialogue context and a psychometric prompt, we approximate the latent affective distribution at each utterance via Monte Carlo sampling under stochastic LLM decoding. The resulting samples are used to estimate the mean affective state and variance, capturing central tendency and ambiguity, respectively. These estimates are subsequently standardized and assembled into longitudinal trajectories for each interlocutor. This pipeline enables uncertainty-aware trajectory representations that support downstream analysis of interpersonal dynamics.

\subsubsection{Probabilistic Affective Perception: Monte Carlo Estimation via Stochastic LLM Decoding}
Our methodology leverages the stochastic properties of Large Language Models (LLMs) to approximate the parameters of the latent affective distribution. We formalize the inference process as a mapping $\mathcal{M}$ from the textual input space to the sentiment trajectory space. The model is conditioned on a domain-specific psychometric prompt $\rho_{\text{emo}}$ and the conversational context window $\mathcal{W} = ((u_t^\text{T}, u_t^\text{S}))_{t=1}^{N}$, where $u_t^\text{T}$ and $u_t^\text{S}$ represent the utterances from the teacher and the student at turn $t$, respectively. The model input $\mathcal{I}$ is constructed via the sequence concatenation of the instruction and the dialogue history: $\mathcal{I} = [\rho_{\text{emo}}; \mathcal{W}].$ To estimate the mean affective state while accounting for linguistic subjectivity, we employ Monte Carlo estimation via stochastic LLM decoding. Specifically, for each turn and each interlocutor $(\cdot) \in \{\text{T}, \text{S}\}$, we perform $K$ independent inference trials under non-zero temperature settings, yielding a set of affective state estimates $\{ \hat{s}_{t,k}^{(\cdot)} \}_{k=1}^K$. This framework assumes that the LLM response distribution, when conditioned on $\rho_{\text{emo}}$, serves as a computational proxy for the inaccessible latent distribution $p(e|u_t)$. Consequently, each trial provides a realization of the latent affective state, denoted as: $\hat{s}_{t,k}^{(\cdot)} \sim \mathcal{M}(\mathcal{I}),$ which facilitates the quantification of affective fluctuations and ambiguity without relying on a single, potentially biased, deterministic output.

\subsubsection{Estimation of Mean Affective State and Ambiguity}
The primary objective of the sampling process is to estimate the fundamental statistical moments of the underlying distribution. In accordance with the framework of \citet{11185123}, the mean of the distribution represents the central emotional tendency, while the variance quantifies the degree of affective ambiguity inherent in the utterance. We define the aggregated raw affective signal $s_t$ as the sample mean derived from the $K$ trials, serving as an empirical estimate of the mean affective state: $s_t^{(\cdot)} = \frac{1}{K} \sum_{k=1}^{K} \hat{s}_{t,k}^{(\cdot)}.$
Furthermore, the variance across these trials, $\sigma^2_t$, is utilized to represent the uncertainty in the model's affective inference: ${\sigma^2_t}^{(\cdot)} = \frac{1}{K-1} \sum_{k=1}^{K} (\hat{s}_{t,k}^{(\cdot)} - s_t^{(\cdot)})^2$.
This variance captures the aleatoric uncertainty associated with linguistic ambiguity.

\subsubsection{Sentiment Scoring Mechanism, Polarity Mapping, and Trajectory Modeling} \label{sec:standardization}
The sentiment quantification process follows the scoring rules defined in $\rho_{\text{emo}}$ (detailed in Appendix~\ref{sec:prompt_appendix}), which utilizes a fixed numerical range from $0$ to $5$ with $0.5$ intervals. This rubric establishes $2.5$ as the neutral baseline for procedural or factual statements; meanwhile, positive emotional cues lead to a numerical decrease toward $0$ and negative cues trigger an increase toward $5$. By employing a non-negative numerical progression instead of signed values during the initial evaluation, the prompt is designed to minimize the risk of logical or sign-related confusion during the model inference process, which serves to enhance the stability and consistency of the generated scores. Under this schema, a raw score of $0.0$ signifies the most positive affective state, such as empathy or successful resolution, whereas $5.0$ represents the most negative state, including explicit conflict or frustration. To align these outputs with affective computing conventions in which positive values represent positive valence, we perform a polarity mapping on the raw results. This transformation maps the averaged raw scores $s_t$ to a centered interval $[-1, 1]$ through the following linear operation: $\tilde{s}_t^{(\cdot)} = 1 - 2\,(s_t^{(\cdot)}/5)$. To maintain statistical consistency, the associated variance is scaled by the square of the mapping coefficient, resulting in $\tilde{\sigma}_t^2 = (\frac{2}{5})^2 \sigma_t^2$. Through this mapping, the value $\tilde{s}_t \to +1$ indicates a strong positive state and $\tilde{s}_t \to -1$ indicates a strong negative state, while $\tilde{s}_t = 0$ remains the neutral point. The transformed variance $\tilde{\sigma}_t^2$ preserves the quantification of affective ambiguity within the standardized space. The final standardized sentiment trajectories are defined as vectors $\tilde{\mathbf{S}}^\text{T}, \tilde{\mathbf{S}}^\text{S} \in [-1, 1]^N$, where each vector $\tilde{\mathbf{S}}^{(\cdot)} = (\tilde{s}_1^{(\cdot)}, \dots, \tilde{s}_N^{(\cdot)})^\top$ captures the continuous evolution of sentiment throughout the dialogue.

\subsection{Interpersonal Affective Dynamics and Interaction Modeling}
\label{sec:xcorr}
Building on the affective trajectories, we model interpersonal dynamics through complementary temporal and trend-based analyses. Sequential alignment and directional influence are quantified via cross-correlation, while long-term affective development is summarized using slope-based indicators. Taken together, these measures serve as interaction indicators that support the interpretation of latent interpersonal behaviors and potential dyadic interaction patterns.

\subsubsection{Sequential Cross-Correlation Analysis}

Our application of cross-correlation for affective time-series analysis is inspired by the methodology from \citet{o2010tweets}, who utilized this technique to link sentiment signals extracted from text streams to longitudinal public opinion data. By adapting this method to the interpersonal dynamics analysis, we can effectively synchronize affective trajectories and identify leading or lagging influences across discrete dialogue turns.

Building on the standardized trajectories $\tilde{\mathbf{S}}^\text{T}$ and $\tilde{\mathbf{S}}^\text{S}$, we employ discrete cross-correlation to quantify the sequential alignment between the interlocutors throughout the dialogue sequence. We define the normalized cross-correlation function (NCCF), $R_{\text{TS}}(L)$, as:
\resizebox{.9\linewidth}{!}{
$\displaystyle
R_{\text{TS}}(L) = \frac{\sum_{t \in \mathcal{T}_L} (\tilde{s}_t^\text{T} - \bar{s}^\text{T}) (\tilde{s}_{t+L}^\text{S} - \bar{s}^\text{S})}
{\sqrt{\sum_{t \in \mathcal{T}_L} (\tilde{s}_t^\text{T} - \bar{s}^\text{T})^2 \sum_{t \in \mathcal{T}_L} (\tilde{s}_{t+L}^\text{S} - \bar{s}^\text{S})^2}}
$
}, where $L \in \mathbb{Z} \cap [-3, 3]$ represents the conversational lag in turns and $\bar{s}^{(\cdot)}$ denotes the sequential mean of the affective scores across the interaction. To ensure mathematical validity over the finite dialogue length $N$, the summation is performed over the set $\mathcal{T}_L = \{t \in \mathbb{Z} \mid 1 \le t \le N, 1 \le t+L \le N\}$, which defines the valid overlapping indices for a given lag $L$. The estimated influence lag $L^\ast = \arg\max_{L} |R_{\text{TS}}(L)|$ identifies the dominant point of sequential phase alignment. Within this framework, $L^\ast \geq 0$ suggests a teacher-leading influence; conversely, $L^\ast < 0$ reflects student-driven adjustment.

\subsubsection{Slope-Based Development Indicator}
To characterize long-term affective development beyond local turn-to-turn alignment, we summarize each standardized sentiment trajectory by its linear trend over time. For each interlocutor $(\cdot)\in\{\text{T},\text{S}\}$, we estimate the slope parameter $\beta_{(\cdot)}$ via least-squares regression over the dialogue turns $t\in\{1,\dots,N\}$, where $N$ denotes the total number of turns considered. In in-text form, the slope is given by
$\beta_{(\cdot)} = \sum_{t=1}^{N} (t-\bar{t})(\tilde{s}_t^{(\cdot)}-\bar{s}^{(\cdot)}) \big/ \sum_{t=1}^{N} (t-\bar{t})^2$,
with $\bar{t}=\frac{1}{N}\sum_{t=1}^{N} t$ and $\bar{s}^{(\cdot)}=\frac{1}{N}\sum_{t=1}^{N}\tilde{s}_t^{(\cdot)}$.
Under the standardized polarity mapping defined in Section~\ref{sec:standardization}, positive values of $\tilde{s}_t$ correspond to positive affect. Accordingly, a positive slope ($\beta_{(\cdot)}>0$) indicates overall emotional amelioration across the interaction, whereas a negative slope ($\beta_{(\cdot)}<0$) reflects affective deterioration. This indicator captures global affective drift and complements the lag-based cross-correlation analysis.

\subsection{From Interpersonal Affective Trajectories to Interaction Insights: Mining Dynamics in Educational Settings}
The transition from interpersonal affective trajectories to interaction insights is achieved by synthesizing the quantitative indicators derived from the turn-based sequence. This analytical framework is designed for broad application across diverse dyadic interactions; however, we utilize educational discourse as a representative case study to illustrate its interpretive capacity. By concurrently analyzing the direction of influence through the estimated influence lag $L^*$, the nature of emotional coupling via $R_{\text{TS}}$, and the long-term progression using the slopes $\beta_{(\cdot)}$, the method identifies specific interaction patterns.

These indicators jointly provide a structured view of affective co-evolution in dialogue. The estimated influence lag captures the temporal ordering of affective influence, distinguishing teacher-led dynamics from student-driven responses. At the same time, trajectory slopes summarize the long-term directional trend of the emotional climate. Together, these signals enable the identification of latent interaction patterns, such as affective synchrony or counterproductive support, without enforcing strict turn-level alignment assumptions. A detailed typology of these joint configurations and their corresponding interpretations in the learning context is summarized in Table~\ref{tab:potential_interpretation}. This joint indicator-based synthesis bridges low-level statistical descriptors estimated during the perception stage with higher-level, interpretable insights into teacher-student affective dynamics \cite{doi:10.1073/pnas.2405460121, linnenbrink2016adaptive, chang2009appraisal, bevilacqua2019brain}.
\begin{table}[t!]
\caption{Joint Indicators-based Interpretation of Teacher-Student Affective Dynamics}
\label{tab:potential_interpretation}
\resizebox{\columnwidth}{!}{%
\begin{tabular}{cccl}
\hline
\textbf{\begin{tabular}[c]{@{}c@{}}Estimated\\Influence Lag ($L^\ast$)\end{tabular}} & \textbf{\begin{tabular}[c]{@{}c@{}}Sign($R_{\text{TS}}(L^*)$)\end{tabular}} & \textbf{Slope Relation}                   & \textbf{Potential Interpretation}                          \\ \hline
$L^\ast > 0$                                                      & $+$                                                           & both $\beta_\text{T}, \beta_\text{S} > 0$ & Effective Scaffolding; teacher-led positive contagion      \\
$L^\ast > 0$                                                      & $+$                                                           & both $\beta_\text{T}, \beta_\text{S} < 0$ & Negative Contagion; teacher-led shared frustration         \\
$L^\ast > 0$                                                      & $-$                                                           & $\beta_\text{T} > 0, \beta_\text{S} < 0$  & Counterproductive Support; teacher regulates student frustration \\ \hline
$L^\ast < 0$                                                      & $+$                                                           & both $\beta_\text{T}, \beta_\text{S} > 0$ & Student-driven success; student motivates teacher          \\
$L^\ast < 0$                                                      & $+$                                                           & both $\beta_\text{T}, \beta_\text{S} < 0$ & Feedback Burnout; student leads shared decline             \\
$L^\ast < 0$                                                      & $-$                                                           & $\beta_\text{S} < 0, \beta_\text{T} > 0$  & Encouragement Escalation; teacher attempts to uplift student  \\ \hline
$L^\ast = 0$                                                      & $+$                                                           & both $\beta_\text{T}, \beta_\text{S} > 0$ & Affective Synchrony; real-time mutual engagement           \\
$L^\ast = 0$                                                      & $+$                                                           & both $\beta_\text{T}, \beta_\text{S} < 0$ & Shared Fatigue; synchronous motivation decline             \\
$L^\ast = 0$                                                      & $-$                                                           & opposite signs                            & Dynamic Compensation; real-time tension balancing          \\ \hline
\end{tabular}%
}
\end{table}

\begin{table*}[]
\caption{Teacher-Student Dialogue Example (\textit{Topic - Personification})}
\label{tab:Personification}
\resizebox{\textwidth}{!}{%
\begin{tabular}{ccclll}
\hline
\multicolumn{4}{l}{\textit{Background Information}}                                                              &                                         &                                                                \\ \hline
\multicolumn{3}{l}{Student Preference:} & Hands-on activities and real-world applications                         & Student Reaction:                        & Becomes rude when disengaged                                   \\
\multicolumn{3}{l}{Teacher Preference:} & Experiential learning and hands-on activities                           & Teacher Reaction:                        & Insists on experiential teaching methods                       \\ \hline
\multicolumn{4}{l}{\textit{Dialogue}}                                                                            &                                         &                                                                \\ \hline
Turn                & Index  & Role    & \multicolumn{3}{l}{Utterance}                                                                                                                                                      \\ \hline
\multirow{2}{*}{0}  & 0      & Teacher & \multicolumn{3}{l}{Today, we're going to learn about personification, a literary device that gives human characteristics to nonhuman things.}                                      \\
                    & 1      & Student & \multicolumn{3}{l}{That sounds boring. Why do we need to know this?}                                                                                                               \\ \hline
\multirow{2}{*}{1}  & 2      & Teacher & \multicolumn{3}{l}{It's a technique that authors use to create vivid and engaging stories. Let's try an activity. Look around the room and find an object that you can personify.} \\
                    & 3      & Student & \multicolumn{3}{l}{This is stupid. I don't want to do this.}                                                                                                                       \\ \hline
\multirow{2}{*}{2}  & 4      & Teacher & \multicolumn{3}{l}{I know it might seem a bit abstract at first, but it's a useful tool to understand. How about we try it with this pencil?}                                      \\
                    & 5      & Student & \multicolumn{3}{l}{Fine, whatever.}                                                                                                                                                \\ \hline
\multirow{2}{*}{3}  & 6      & Teacher & \multicolumn{3}{l}{Imagine the pencil as a person. What kind of personality would it have?}                                                                                        \\
                    & 7      & Student & \multicolumn{3}{l}{I don't know. It's just a pencil.}                                                                                                                              \\ \hline
\multirow{2}{*}{4}  & 8      & Teacher & \multicolumn{3}{l}{Think about its qualities. It's sharp, it can write, it's often used to express ideas.}                                                                         \\
                    & 9      & Student & \multicolumn{3}{l}{Maybe it could be a shy writer who's afraid to share its thoughts?}                                                                                             \\ \hline
\multirow{2}{*}{5}  & 10     & Teacher & \multicolumn{3}{l}{That's a great example! You're starting to get the hang of it. Now, let's write a short story where the pencil is the main character.}                          \\
                    & 11     & Student & \multicolumn{3}{l}{Okay, this is actually kind of fun.}                                                                                                                            \\ \hline
\multirow{2}{*}{6}  & 12     & Teacher & \multicolumn{3}{l}{I'm glad you're enjoying it. Remember, personification helps us relate to objects and stories in a new way.}                                                    \\
                    & 13     & Student & \multicolumn{3}{l}{I think I get it now. Thanks for showing me this way.}                                                                                                          \\ \hline
\multirow{2}{*}{7}  & 14     & Teacher & \multicolumn{3}{l}{You're welcome. I'm happy you found it engaging.}                                                                                                               \\
                    & 15     & Student & \multicolumn{3}{l}{{[}End of conversation{]}}                                                                                                                                      \\ \hline
\end{tabular}%
}
\end{table*}

\begin{figure*}[t]
\centering
\begin{minipage}{0.32\textwidth}
    \centering
    \includegraphics[width=\linewidth]{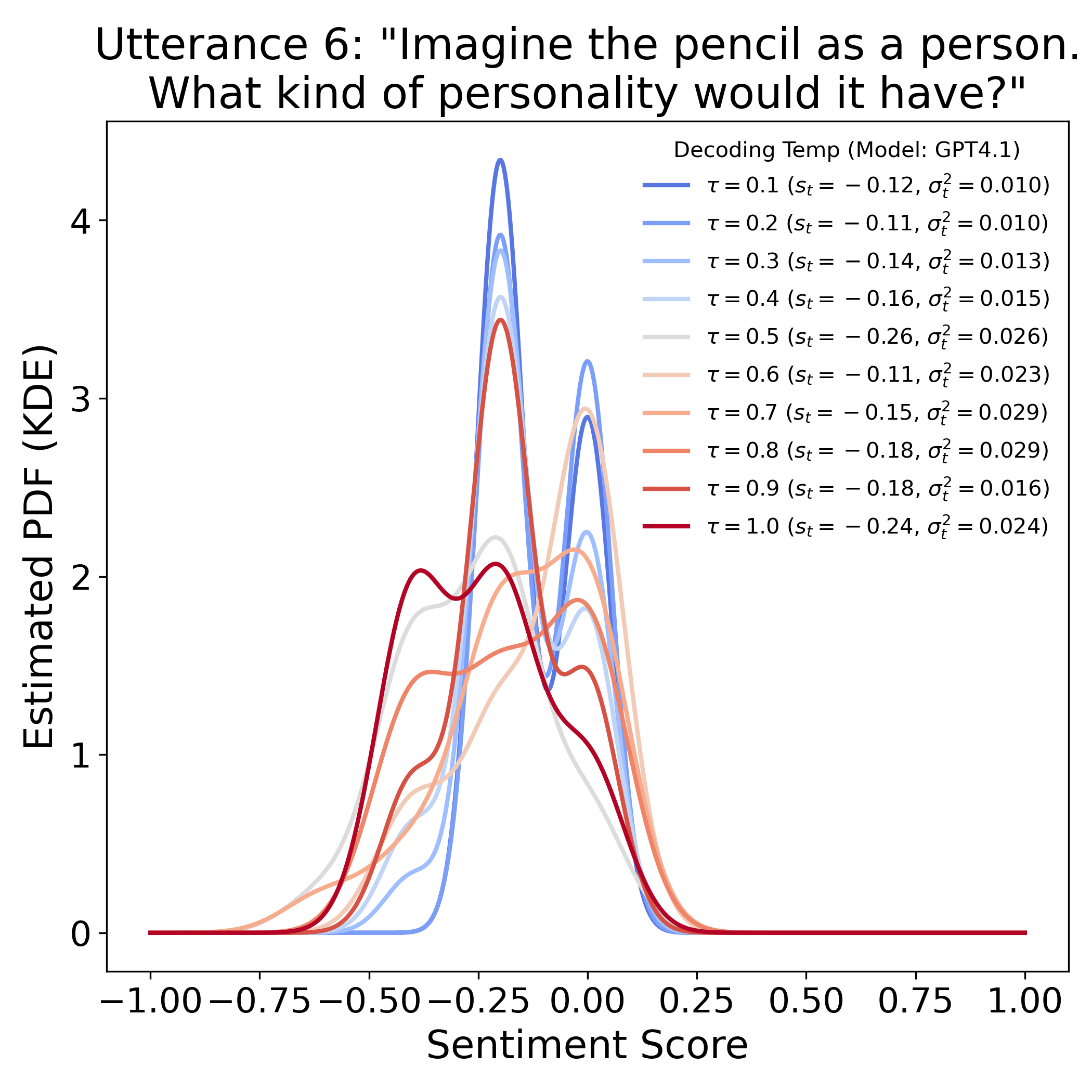}
    \subcaption{}
    \label{fig:sensitivity_distribution}
\end{minipage}
\hfill
\begin{minipage}{0.32\textwidth}
    \centering
    \includegraphics[width=\linewidth]{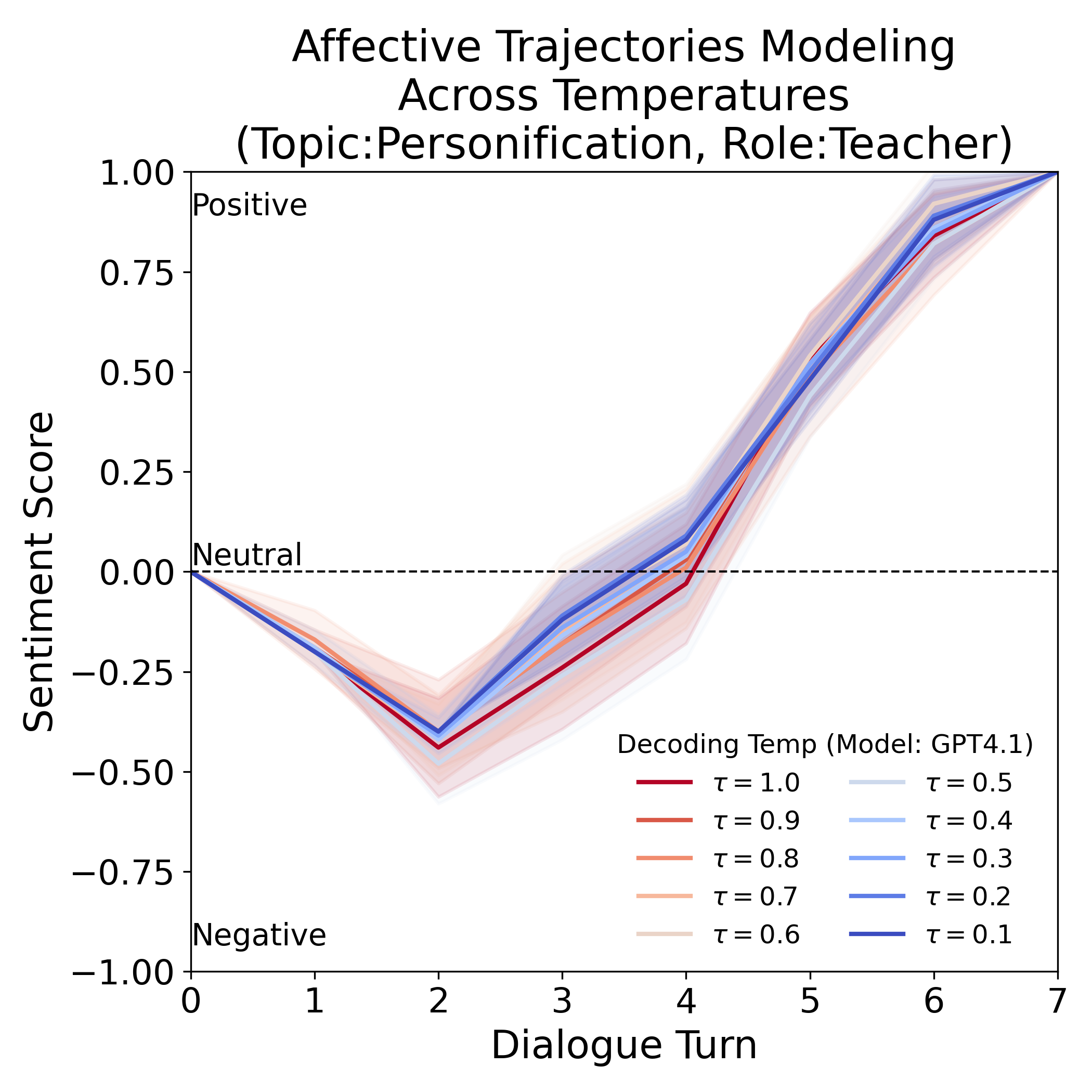}
    \subcaption{}
    \label{fig:sensitivity_trajector_xtemp}
\end{minipage}
\hfill
\begin{minipage}{0.32\textwidth}
    \centering
    \includegraphics[width=\linewidth]{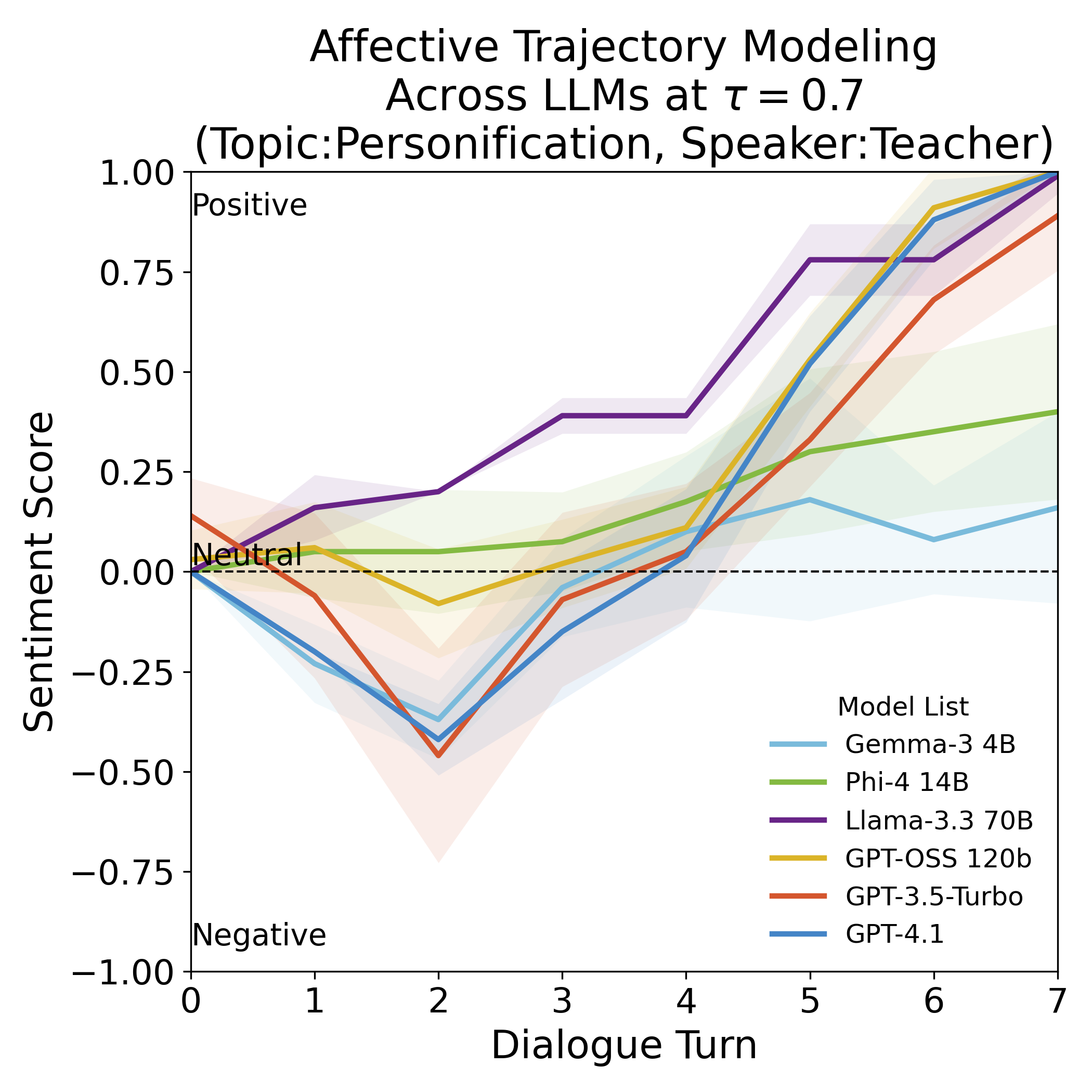}
    \subcaption{}
    \label{fig:xmodel_analysis}
\end{minipage}
\hfill
\caption{Sensitivity analysis of probabilistic affective estimation and trajectory stability. 
(a) Probabilistic sentiment score distributions for Utterance~6 in the \textit{Personification} topic across decoding temperatures ($\tau \in [0.1, 1.0]$), estimated via Kernel Density Estimation (KDE). 
(b) Convergence of teacher mean affective trajectories for the \textit{Personification} topic, demonstrating the stability of the estimated mean affective state across the temperature range. 
(c) Comparative affective trajectory modeling across six LLMs at a fixed decoding temperature ($\tau=0.7$). \\\textbf{*Note:} In (b) and (c), shaded uncertainty bands denote $\pm 1$ standard deviation ($\pm\sigma$), representing the quantified affective ambiguity in the models’ perceptions.}

\label{fig:temp_sensitivity}
\end{figure*}

\section{Experiment results and Discussion}
This section evaluates the robustness of LLM-MC-Affect under varying temperatures and models, and presents case studies illustrating its ability to capture interpersonal dynamics in educational dialogues.
\subsection{Experimental Setup}
We evaluate our approach using Google’s Education Dialogue Dataset \cite{10.5555/3737916.3741695}, a recently introduced synthetic dataset for modeling multi-turn instructional interactions. The use of a simulated dialogue environment is motivated by well-established logistical and ethical constraints in educational research \cite{nebeker2017ethical}. Collecting data from real classroom settings typically requires extensive Institutional Review Board (IRB) approval and privacy concerns associated with student information. Accordingly, this study is positioned as an empirical pilot that validates the proposed affective modeling framework under controlled conditions before deployment in naturalistic educational environments. Importantly, recent studies have demonstrated that LLMs can effectively simulate complex human behaviors and social dynamics, and are increasingly adopted in computational social science and behavioral research \cite{doi:10.1073/pnas.2518075122, 10.1145/3586183.3606763, doi:10.1073/pnas.2405460121}. Such simulation-based datasets enable systematic control over participant personas and interaction variables, which is difficult to achieve in real-world classroom studies.

For evaluation, we access \texttt{GPT-4.1} (snapshot 2025-04-14) and \texttt{GPT-3.5-Turbo} (snapshot 2024-01-25) via the OpenAI API \cite{achiam2023gpt, ouyang2022training}, while open-source models, including \texttt{Gemma 3 4B}, \texttt{Llama 3.3 70B}, \texttt{Phi 4 14B} and \texttt{GPT-OSS 120B}, are deployed through the AI Verde service \cite{team2025gemma, dubey2024llama, abdin2024phi, agarwal2025gpt, mithun2025ai}. All models are evaluated in their original, non-fine-tuned configurations to ensure that the results reflect their inherent zero-shot affective perception capabilities. To ensure empirical consistency and reproducibility, a unified scoring rubric ($\rho_{\text{emo}}$) is applied across all experiments. For each dialogue turn, we conduct $K=20$ independent Monte Carlo trials to estimate parameters of the latent affective distribution. In the main text, we present a detailed analysis of a representative case study, \textit{Personification} (detailed in Table \ref{tab:Personification}), to illustrate how the proposed quantitative indicators yield pedagogically meaningful insights. Additional case studies are reported in Appendix \ref{sec:app_additional}.

\subsection{Sensitivity Analysis of Temperature on Probabilistic Affective Perception}
We treat the decoding temperature as a controllable stochasticity parameter and perform a sensitivity analysis using \texttt{GPT-4.1} as the primary model \cite{liu2024emollms} by examining how temperature variations affect the estimated affective distributions and trajectory stability with $\rho_{\text{emo}}$. This approach is grounded in the statistical formulation that characterizes the emotion of an utterance not as a deterministic point estimate, but as a latent probability distribution $p(e|u_t)$. By leveraging the stochastic properties of LLMs at non-zero temperature, we can approximate this latent distribution via Monte Carlo sampling.
\subsubsection{Impact on Distributional Characteristics}
The decoding temperature $\tau$ serves as a critical modulator of the model's exploratory behavior during sentiment quantification. As illustrated in Figure \ref{fig:sensitivity_distribution} (generated via \texttt{GPT-4.1}), the Kernel Density Estimation (KDE) for Utterance 6 - \textit{"Imagine the pencil as a person. What kind of personality would it have?"}, lower temperature settings reveal a complex, bimodal perception of pedagogical intent. At $\tau = 0.1$, the distribution is characterized by two distinct, sharp peaks centered near $-0.20$ and $0.00$, with a recorded sample mean $\tilde{s}_t = -0.12$ and a minimal variance $\tilde{\sigma}_t^2 = 0.010$. These sharp peaks indicate a pseudo-deterministic interpretation in which the model identifies two divergent yet high-confidence affective interpretations of the prompt. As the temperature approaches 1.0, these distinct peaks merge into a broader, more continuous distribution. Higher temperature does not necessarily imply a more accurate affective judgment; rather, it enables the LLM to explore a broader set of plausible interpretations of the utterance, thereby making potential affective ambiguity more visible in the resulting sample distribution. While the variance captures the aleatoric uncertainty associated with linguistic ambiguity, the distribution exhibits reduced concentration, with the variance increasing to $\tilde{\sigma}_t^2 = 0.024$ at $\tau = 1.0$. This behavior aligns with our statistical formulation, where ambiguity is treated as an essential emotional nuance rather than mere noise. By leveraging higher stochasticity, this method reveals the hidden density of the affective variable, providing a quantitative proxy for interpretative subjectivity that would be lost in deterministic settings. For completeness, the utterance-level mean affective states and variances across all temperature settings are reported in Appendix~\ref{sec:app_q_detail}.

\subsubsection{Robustness and Reliability of the Mean Affective State}
A fundamental finding of this sensitivity analysis is the robustness of the mean affective state despite significant shifts in distribution. While the variance $\sigma_t^2$ effectively doubles as $\tau$ scales from $0.1$ to $1.0$, the sample mean $s_t$ remains remarkably stable and concentrated within a narrow band. For Utterance 6 in the personification topic dialogue, the mean fluctuates only marginally between $-0.11$ and $-0.26$ across the entire temperature evaluation range. The reliability of these mean estimates is further demonstrated at the level of affective trajectory modeling. In the Figure \ref{fig:sensitivity_trajector_xtemp} (generated via \texttt{GPT-4.1}) for the Personification topic, the teacher's affective trajectories across all $\tau$ values exhibit a high degree of convergence. All plotted lines consistently track the same emotional progression, initially dipping toward negative valence at Turn 2 before rising steadily to a strong positive state by Turn 7. This convergence confirms that the framework can filter out stochastic sampling noise while preserving the underlying affective signal. This stability ensures that the derived interpersonal metrics in downstream analyses, such as the estimated influence lag ($L^{*}$) and Cross-Correlation ($R_{TS}$), remain reliable indicators of interaction dynamics despite minor sampling variations.

\subsubsection{Cross-Model Behavior Evaluation}
To assess cross-model behavior, we compare affective trajectories produced by six LLMs on the Personification dialogue at a fixed decoding temperature of $\tau = 0.7$. This setting balances mean stability and perceptual ambiguity, enabling meaningful Monte Carlo variance estimation without degrading trajectory convergence. All models share the same psychometric prompt ($\rho_{\text{emo}}$) and operate in a zero-shot configuration. Results are reported in Figure~\ref{fig:xmodel_analysis}. The result shows that \texttt{GPT-4.1} and \texttt{GPT-3.5-Turbo} exhibit the strongest structural consistency, both capturing the characteristic V-shaped affective progression with an early negative dip followed by recovery. However, \texttt{GPT-3.5-Turbo} exhibits substantially higher affective ambiguity, as indicated by a wider uncertainty band ($\pm\sigma$), suggesting greater stochastic volatility and a less confident estimate of the central affective state. In contrast, applying the same analysis to open-source models exposes pronounced divergences in affective resolution, revealing systematic biases that limit their ability to capture fine-grained emotional dynamics:
\begin{itemize}
    \item \textbf{Extreme Positivity Bias (\texttt{Llama 3.3 70B}):} The Llama 3.3 model demonstrates a near-total failure in identifying the negative emotional shift present in the early dialogue. Its trajectory follows a continuous, near-linear upward trend from the neutral baseline, failing to detect any dip at Turn 2. The Llama 3.3’s insensitivity to negative pedagogical cues is likely a result of alignment for harmlessness, a process designed to ensure models produce helpful and harmless responses \cite{dubey2024llama, mozikov2024eai}. 
    \item \textbf{Performance Gaps and Conservative Biases (\texttt{GPT-OSS 120B}, \texttt{Gemma 3 4B}, and \texttt{Phi-4 14B})}: 
    Across models outside the GPT family, a consistent pattern of conservative affective estimation is observed, manifesting as systematic overestimation of affective states during negative pedagogical turns. Despite its large parameter scale, \texttt{GPT-OSS 120B} exhibits a markedly shallow emotional dip during the Turn 1-2 transition, substantially weaker than that of the GPT series. This conservative response is consistent with a cognitive appraisal bias \cite{tak2025aware}. Similar but more pronounced limitations appear in smaller architectures \cite{liu2024emollms}. \texttt{Phi-4 14B} displays a mild positive bias; although its uncertainty band partially overlaps the negative region at Turn~2, its mean trajectory remains elevated and its subsequent recovery is capped at approximately 0.40. In addition, \texttt{Gemma 3 4B} exhibits a recovery bottleneck: while it detects the initial dip, it fails to model the rebound phase, with its trajectory plateauing near the neutral baseline ($\sim$0.15). Collectively, these patterns indicate that both large and small non-GPT models tend to underestimate affective turning points, albeit for different architectural or alignment-related reasons.
\end{itemize}

\begin{figure}[t!]
    \centering
    \begin{minipage}{0.8\linewidth}
        \centering
        \includegraphics[width=\linewidth]{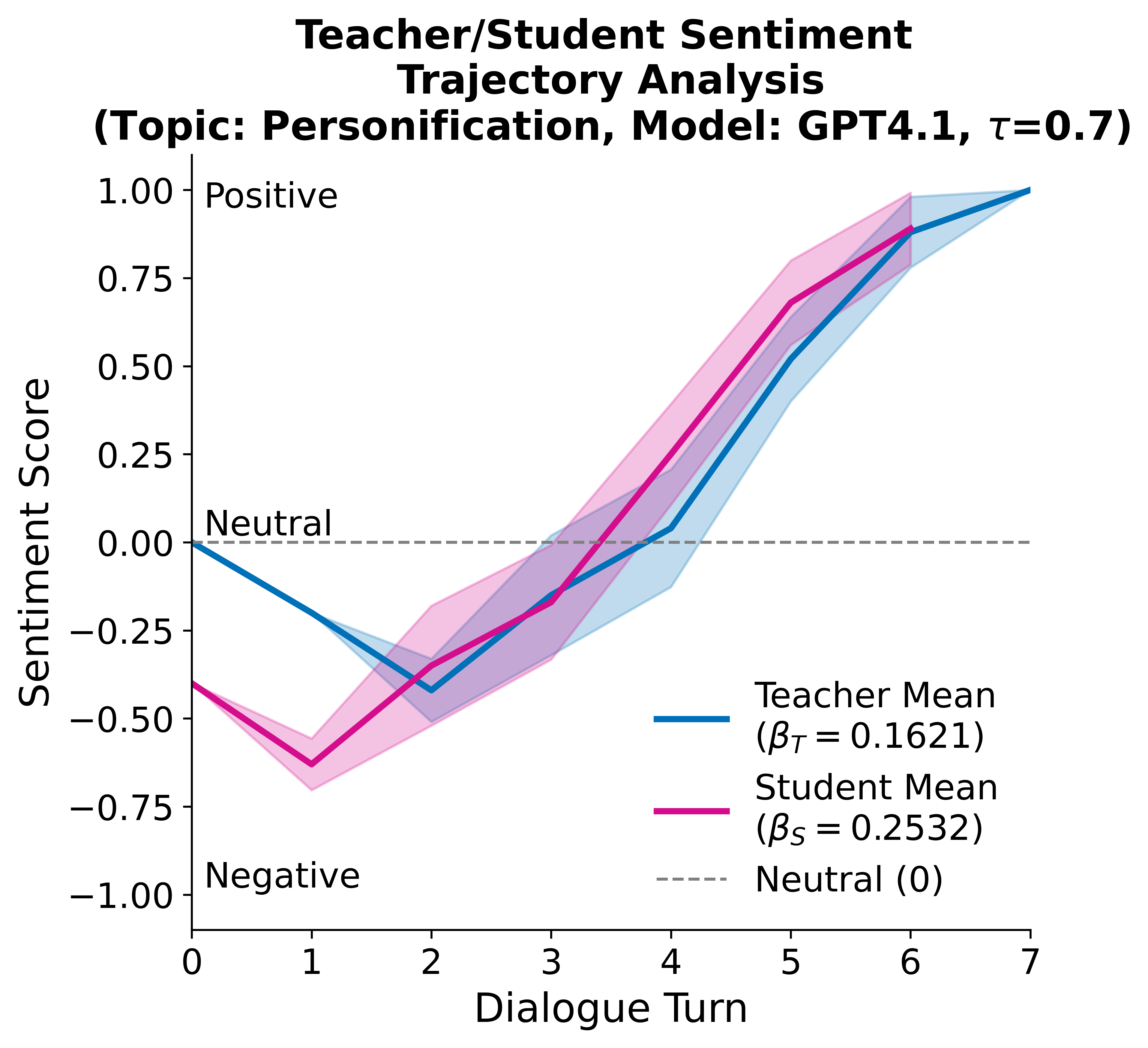}
        \subcaption{}
        \label{fig:example_a}
    \end{minipage}
    \hfill
    \begin{minipage}{0.8\linewidth}
        \centering
        \includegraphics[width=\linewidth]{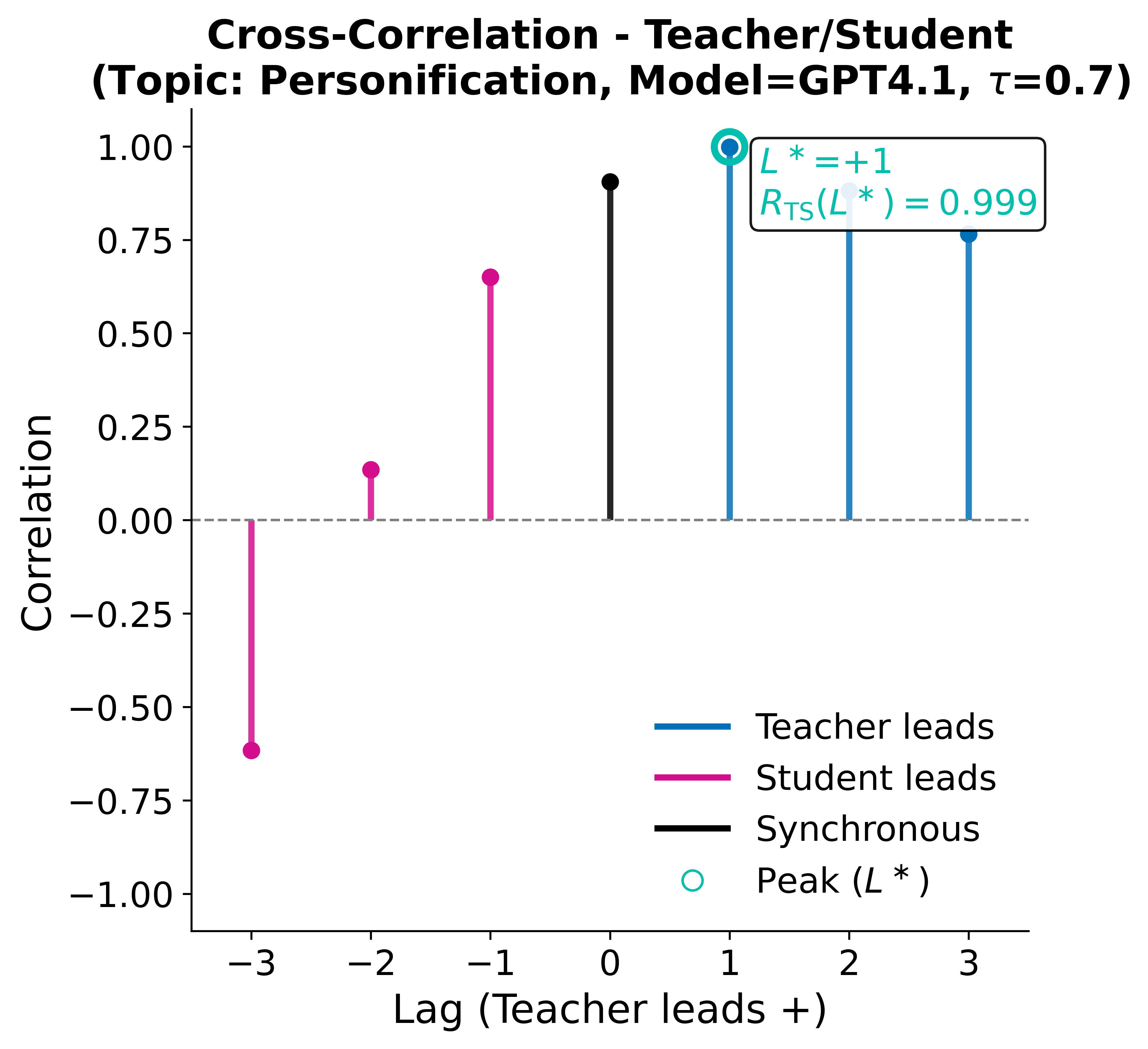}
        \subcaption{}
        \label{fig:example_b}
    \end{minipage}
    \caption{Illustration of affective trajectory modeling and interpersonal dynamics analysis for the \textit{Personification} dialogue using LLM-MC-Affect (\texttt{GPT-4.1}, $\tau = 0.7$). (a) Estimated teacher and student affective sentiment trajectories modeling, where solid lines denote mean affective states and shaded regions indicate uncertainty bands ($\pm\sigma$). Slopes: $\beta_T = 0.1621, \beta_S = 0.2532$ (b) NCCF analysis between teacher and student trajectories across conversational lags ($L^* = +1$, $R_{TS}(L^*) = 0.999$).}
    \label{fig:interpersonal_dynamics}
\end{figure}

\subsection{Interpreting Interpersonal Dynamics and Interaction Insights}

Based on our cross-model stability analysis, \texttt{GPT-4.1} demonstrates a superior capacity to effectively reflect subtle fluctuations in dialogue emotion. It maintains high resolution in affective trajectory modeling, even in scenarios where other architectures exhibit conservative patterns or systematic biases. This empirical observation aligns with prior work, such as \citet{liu2024emollms}, which demonstrates that \texttt{GPT-4} outperforms other models on complex emotional recognition tasks. Consequently, we proceed with our interactional analysis using the affective scores derived from \texttt{GPT-4.1} at a fixed decoding temperature of $\tau = 0.7$. This temperature setting serves as a practical balance point that preserves the central emotional tendency while providing sufficient stochastic variance to represent latent affective ambiguity. Using this representation, we examine the interpersonal dynamics and emotional shifts within the instructional context, with the resulting affective trajectories and NCCF analysis shown in Figure~\ref{fig:interpersonal_dynamics}. Synthesizing the sequential and directional metrics derived from our framework, we categorize the interaction for the \textit{"Personification"} topic as \textbf{Effective Scaffolding}. This classification, based on the definitions in Table \ref{tab:potential_interpretation}, reflects a scenario where the teacher actively drives a positive emotional climate. NCCF analysis reveals that the teacher acts as the leading signal, with an estimated influence delay of $L^* = +1$ yielding a near-perfect correlation of $R_{TS} = 0.999$. This indicates that the student's affective state is highly predictable based on the teacher's emotional output from the preceding turn, confirming the teacher's role as the primary driver of interpersonal dynamics.

\subsubsection{Alignment with Persona Preferences and Reactions in Setting of \textit{Personification} Topic}
The quantitative metrics align remarkably well with the specific persona traits and reaction settings defined for this interaction:
\begin{itemize}
    \item \textbf{Teacher - Insistence on Experiential Methods}: The teacher's preference for \textit{experiential learning} and their reaction setting of \textit{insisting on experiential methods} are reflected in the proactive leading role ($L^* = +1$). This sequential dominance shows the teacher successfully steering the dialogue toward their preferred instructional style.
    \item \textbf{Student - Hands-on Preference and Rude Reactions}: The student's preference for \textit{hands-on activities} and their tendency to \textit{become rude when disengaged} serve as critical validation points. The near-perfect correlation and the positive student slope confirm that the student's preferences were met through the personification exercise, effectively preventing a rude reaction.
\end{itemize}

\section{Conclusion}
This work presented LLM-MC-Affect, a probabilistic and ambiguity-aware framework for affective perception in conversational text that models utterance-level affect as a latent affective distribution rather than a deterministic label. By combining stochastic LLM decoding with Monte Carlo sampling, the framework estimates both central affective tendency and perceptual ambiguity in natural language. This approach enables the construction of stable, uncertainty-aware affective trajectories that remain robust under decoding stochasticity and avoid reliance on supervised annotation or biometric sensing.

Building on these trajectories, we introduce an interaction-level analysis layer that examines affective dynamics through a signal-analytic perspective, integrating sequential cross-correlation, influence lag estimation, and slope-based trend indicators to characterize sequential alignment, directional influence, and long-term affective progression between interlocutors. Empirical results demonstrate that the proposed framework reliably preserves trajectory stability under decoding stochasticity while explicitly exposing affective ambiguity, and that, when instantiated with affect-sensitive LLMs, it yields interpretable interaction patterns aligned with pedagogical structure and dialogue context. By bridging probabilistic affect modeling with interaction-level explanation, LLM-MC-Affect provides a scalable and deployable pathway toward affect-aware analytics for adaptive systems in educational and broader interpersonal settings.

\section*{Limitations}
The proposed framework interprets variance observed across stochastic LLM decoding runs as a proxy for affective uncertainty. While the use of a highly controlled simulated instructional dataset enables systematic control of dialogue structure and decoding conditions, it does not entirely eliminate uncertainty introduced by the language models themselves. In practice, the observed variance may conflate multiple sources of uncertainty, including epistemic effects arising from model biases, prompt sensitivity, and decoding stochasticity, as well as any ambiguity inherent in the input utterances. Given the opaque internal representations of contemporary LLMs, these factors cannot be isolated or measured directly. Consequently, the Monte Carlo variance in this study is treated as an estimated signal rather than as a definitive measure of human-perceived affective ambiguity. It is designed to support, rather than replace, qualitative interpretation of the resulting scores and trajectories.

This work is evaluated on a simulated educational dialogue dataset in which interactions are generated under controlled conditions. While this setting enables systematic and replicable analysis, it does not fully reflect the complexity and variability of real-world human interactions. In practical deployment, especially in educational settings, biases in large language models may lead to overly optimistic or inaccurate interpretations of students’ affective states. Such misestimations could reduce the effectiveness of the proposed framework and may disadvantage vulnerable or underrepresented students by failing to identify their actual needs. Accordingly, future work will focus on evaluation in real educational environments, systematic bias auditing across diverse learner populations, and the incorporation of human oversight mechanisms to improve reliability and fairness in deployment.

Moreover, cross-correlation should not be interpreted as evidence of a causal relationship between the teacher’s and student’s affective trajectories. Consistent with prior time-series sentiment studies, correlation-based lag indicators are intended to capture patterns of sequential alignment rather than causal mechanisms. This limitation is further affected by the relatively short length of the evaluated dialogues, making the statistical robustness of the results and their generalizability to longer interactions uncertain. For longer conversations, future work may incorporate sliding-window or weighted sequential analyses to capture locally stable affective dynamics while mitigating the influence of global correlations.

Finally, the framework incurs substantial computational cost due to repeated stochastic decoding. The present study benchmarks performance using general-purpose LLMs to assess feasibility across model families; however, this choice may not be optimal in terms of efficiency. Future work may explore task-specific affective models or hybrid architectures to achieve a more favorable trade-off between affective sensitivity, interpretability, and computational efficiency.

\section*{Acknowledgments}
This work was partially supported by the National Science
Foundation (NSF) under research project 2335046, and the OpenAI Researcher Access Program 0000011862. We gratefully recognize the AI Verde team at the University of Arizona Data Science Institute for their support and resources, which enabled the use of open-source LLMs in this study.
\bibliography{custom}

@inproceedings{o2010tweets,
  title={From tweets to polls: Linking text sentiment to public opinion time series},
  author={O'Connor, Brendan and Balasubramanyan, Ramnath and Routledge, Bryan and Smith, Noah},
  booktitle={Proceedings of the international AAAI conference on web and social media},
  volume={4},
  number={1},
  pages={122--129},
  year={2010}
}

@article{10.1093/scan/nsab103,
    author = {Nguyen, Mai and Chang, Ashley and Micciche, Emily and Meshulam, Meir and Nastase, Samuel A and Hasson, Uri},
    title = {Teacher–student neural coupling during teaching and learning},
    journal = {Social Cognitive and Affective Neuroscience},
    volume = {17},
    number = {4},
    pages = {367-376},
    year = {2021},
    month = {08},
    issn = {1749-5016},
    doi = {10.1093/scan/nsab103},
    url = {https://doi.org/10.1093/scan/nsab103},
    eprint = {https://academic.oup.com/scan/article-pdf/17/4/367/43222022/nsab103.pdf},
}

@article{qi2024dyadic,
  title={How dyadic emotional transmission shapes teacher-student relationship: Effects of emotional convergence on cohesion in teacher-student interaction},
  author={Qi, Huizi and Zhang, Yulin and Dong, Kaixi and Zhao, Guozhen},
  journal={Current Psychology},
  volume={43},
  number={28},
  pages={23469--23483},
  year={2024},
  publisher={Springer}
}

@ARTICLE{11185123,
  author={Wu, Jingyao and Dang, Ting and Sethu, Vidhyasaharan and Ambikairajah, Eliathamby},
  journal={IEEE Transactions on Affective Computing}, 
  title={How many raters do we need? Analyses of uncertainty in estimating ambiguity-aware emotion labels}, 
  year={2025},
  volume={},
  number={},
  pages={1-15},
  doi={10.1109/TAFFC.2025.3616071}}

@inproceedings{10.5555/3737916.3741695,
author = {Shani, Lior and Rosenberg, Aviv and Cassel, Asaf and Lang, Oran and Calandriello, Daniele and Zipori, Avital and Noga, Hila and Keller, Orgad and Piot, Bilal and Szpektor, Idan and Hassidim, Avinatan and Matias, Yossi and Munos, R\'{e}mi},
title = {Multi-turn reinforcement learning from preference human feedback},
year = {2024},
isbn = {9798331314385},
publisher = {Curran Associates Inc.},
address = {Red Hook, NY, USA},
booktitle = {Proceedings of the 38th International Conference on Neural Information Processing Systems},
articleno = {3779},
numpages = {41},
location = {Vancouver, BC, Canada},
series = {NIPS '24}
}

@article{
doi:10.1073/pnas.2518075122,
author = {Sean J. Westwood },
title = {The potential existential threat of large language models to online survey research},
journal = {Proceedings of the National Academy of Sciences},
volume = {122},
number = {47},
pages = {e2518075122},
year = {2025},
doi = {10.1073/pnas.2518075122},
URL = {https://www.pnas.org/doi/abs/10.1073/pnas.2518075122},
eprint = {https://www.pnas.org/doi/pdf/10.1073/pnas.2518075122}}

@inproceedings{10.1145/3586183.3606763,
author = {Park, Joon Sung and O'Brien, Joseph and Cai, Carrie Jun and Morris, Meredith Ringel and Liang, Percy and Bernstein, Michael S.},
title = {Generative Agents: Interactive Simulacra of Human Behavior},
year = {2023},
isbn = {9798400701320},
publisher = {Association for Computing Machinery},
address = {New York, NY, USA},
url = {https://doi.org/10.1145/3586183.3606763},
doi = {10.1145/3586183.3606763},
booktitle = {Proceedings of the 36th Annual ACM Symposium on User Interface Software and Technology},
articleno = {2},
numpages = {22},
location = {San Francisco, CA, USA},
series = {UIST '23}
}

@article{mithun2025ai,
  title={AI-VERDE: A Gateway for Egalitarian Access to Large Language Model-Based Resources For Educational Institutions},
  author={Mithun, Paul and Noriega-Atala, Enrique and Merchant, Nirav and Skidmore, Edwin},
  journal={arXiv preprint arXiv:2502.09651},
  year={2025}
}

@article{team2025gemma,
  title={Gemma 3 technical report},
  author={Team, Gemma and Kamath, Aishwarya and Ferret, Johan and Pathak, Shreya and Vieillard, Nino and Merhej, Ramona and Perrin, Sarah and Matejovicova, Tatiana and Ram{\'e}, Alexandre and Rivi{\`e}re, Morgane and others},
  journal={arXiv preprint arXiv:2503.19786},
  year={2025}
}

@article{abdin2024phi,
  title={Phi-4 technical report},
  author={Abdin, Marah and Aneja, Jyoti and Behl, Harkirat and Bubeck, S{\'e}bastien and Eldan, Ronen and Gunasekar, Suriya and Harrison, Michael and Hewett, Russell J and Javaheripi, Mojan and Kauffmann, Piero and others},
  journal={arXiv preprint arXiv:2412.08905},
  year={2024}
}

@article{nebeker2017ethical,
  title={Ethical and regulatory challenges of research using pervasive sensing and other emerging technologies: IRB perspectives},
  author={Nebeker, Camille and Harlow, John and Espinoza Giacinto, Rebeca and Orozco-Linares, Rubi and Bloss, Cinnamon S and Weibel, Nadir},
  journal={AJOB empirical bioethics},
  volume={8},
  number={4},
  pages={266--276},
  year={2017},
  publisher={Taylor \& Francis}
}

@article{dubey2024llama,
  title={The llama 3 herd of models},
  author={Dubey, Abhimanyu and Jauhri, Abhinav and Pandey, Abhinav and Kadian, Abhishek and Al-Dahle, Ahmad and Letman, Aiesha and Mathur, Akhil and Schelten, Alan and Yang, Amy and Fan, Angela and others},
  journal={arXiv preprint arXiv:2407.21783},
  year={2024}
}

@article{agarwal2025gpt,
  title={gpt-oss-120b \& gpt-oss-20b model card},
  author={Agarwal, Sandhini and Ahmad, Lama and Ai, Jason and Altman, Sam and Applebaum, Andy and Arbus, Edwin and Arora, Rahul K and Bai, Yu and Baker, Bowen and Bao, Haiming and others},
  journal={arXiv preprint arXiv:2508.10925},
  year={2025}
}

@article{achiam2023gpt,
  title={GPT-4 technical report},
  author={Achiam, Josh and Adler, Steven and Agarwal, Sandhini and Ahmad, Lama and Akkaya, Ilge and Aleman, Florencia Leoni and Almeida, Diogo and Altenschmidt, Janko and Altman, Sam and Anadkat, Shyamal and others},
  journal={arXiv preprint arXiv:2303.08774},
  year={2023}
}

@article{ouyang2022training,
  title={Training language models to follow instructions with human feedback},
  author={Ouyang, Long and Wu, Jeffrey and Jiang, Xu and Almeida, Diogo and Wainwright, Carroll and Mishkin, Pamela and Zhang, Chong and Agarwal, Sandhini and Slama, Katarina and Ray, Alex and others},
  journal={Advances in neural information processing systems},
  volume={35},
  pages={27730--27744},
  year={2022}
}

@article{mozikov2024eai,
  title={EAI: Emotional decision-making of LLMs in strategic games and ethical dilemmas},
  author={Mozikov, Mikhail and Severin, Nikita and Bodishtianu, Valeria and Glushanina, Maria and Nasonov, Ivan and Orekhov, Daniil and Vladislav, Pekhotin and Makovetskiy, Ivan and Baklashkin, Mikhail and Lavrentyev, Vasily and others},
  journal={Advances in Neural Information Processing Systems},
  volume={37},
  pages={53969--54002},
  year={2024}
}

@article{tak2025aware,
  title={Aware yet biased: Investigating emotional reasoning and appraisal bias in large language models},
  author={Tak, Ala N and Gratch, Jonathan and Scherer, Klaus R},
  journal={IEEE Transactions on Affective Computing},
  year={2025},
  publisher={IEEE}
}

@inproceedings{liu2024emollms,
  title={Emollms: A series of emotional large language models and annotation tools for comprehensive affective analysis},
  author={Liu, Zhiwei and Yang, Kailai and Xie, Qianqian and Zhang, Tianlin and Ananiadou, Sophia},
  booktitle={Proceedings of the 30th ACM SIGKDD Conference on Knowledge Discovery and Data Mining},
  pages={5487--5496},
  year={2024}
}

@article{
doi:10.1073/pnas.2405460121,
author = {Michal Kosinski },
title = {Evaluating large language models in theory of mind tasks},
journal = {Proceedings of the National Academy of Sciences},
volume = {121},
number = {45},
pages = {e2405460121},
year = {2024},
doi = {10.1073/pnas.2405460121},
URL = {https://www.pnas.org/doi/abs/10.1073/pnas.2405460121},
eprint = {https://www.pnas.org/doi/pdf/10.1073/pnas.2405460121}}

@article{linnenbrink2016adaptive,
  title={Adaptive motivation and emotion in education: Research and principles for instructional design},
  author={Linnenbrink-Garcia, Lisa and Patall, Erika A and Pekrun, Reinhard},
  journal={Policy Insights from the Behavioral and Brain Sciences},
  volume={3},
  number={2},
  pages={228--236},
  year={2016},
  publisher={Sage Publications Sage CA: Los Angeles, CA}
}

@article{chang2009appraisal,
  title={An appraisal perspective of teacher burnout: Examining the emotional work of teachers},
  author={Chang, Mei-Lin},
  journal={Educational psychology review},
  volume={21},
  number={3},
  pages={193--218},
  year={2009},
  publisher={Springer}
}

@article{bevilacqua2019brain,
  title={Brain-to-brain synchrony and learning outcomes vary by student--teacher dynamics: Evidence from a real-world classroom electroencephalography study},
  author={Bevilacqua, Dana and Davidesco, Ido and Wan, Lu and Chaloner, Kim and Rowland, Jess and Ding, Mingzhou and Poeppel, David and Dikker, Suzanne},
  journal={Journal of cognitive neuroscience},
  volume={31},
  number={3},
  pages={401--411},
  year={2019},
  publisher={MIT Press One Rogers Street, Cambridge, MA 02142-1209, USA journals-info~…}
}

@article{davani2022dealing,
  title={Dealing with disagreements: Looking beyond the majority vote in subjective annotations},
  author={Davani, Aida Mostafazadeh and D{\'\i}az, Mark and Prabhakaran, Vinodkumar},
  journal={Transactions of the Association for Computational Linguistics},
  volume={10},
  pages={92--110},
  year={2022},
  publisher={MIT Press One Rogers Street, Cambridge, MA 02142-1209, USA journals-info~…}
}

@inproceedings{
kuhn2023semantic,
title={Semantic Uncertainty: Linguistic Invariances for Uncertainty Estimation in Natural Language Generation},
author={Lorenz Kuhn and Yarin Gal and Sebastian Farquhar},
booktitle={The Eleventh International Conference on Learning Representations },
year={2023},
url={https://openreview.net/forum?id=VD-AYtP0dve}
}

@article{lin2025personalized,
  title={Personalized Education with Generative AI and Digital Twins: VR, RAG, and Zero-Shot Sentiment Analysis for Industry 4.0 Workforce Development},
  author={Lin, Yu-Zheng and Petal, Karan and Alhamadah, Ahmed H and Ghimire, Sujan and Redondo, Matthew William and Corona, David Rafael Vidal and Pacheco, Jesus and Salehi, Soheil and Satam, Pratik},
  journal={arXiv preprint arXiv:2502.14080},
  year={2025}
}

@article{zhang2025dialoguellm,
  title={Dialoguellm: Context and emotion knowledge-tuned large language models for emotion recognition in conversations},
  author={Zhang, Yazhou and Wang, Mengyao and Wu, Youxi and Tiwari, Prayag and Li, Qiuchi and Wang, Benyou and Qin, Jing},
  journal={Neural Networks},
  pages={107901},
  year={2025},
  publisher={Elsevier}
}

@article{liu2024text,
  title={Text mining and sentiment analysis: A new lens to explore the emotion dynamics of mother-child interactions},
  author={Liu, Chao and Chen, Charis},
  journal={Social Development},
  volume={33},
  number={3},
  pages={e12733},
  year={2024},
  publisher={Wiley Online Library}
}

@inproceedings{renze2024effect,
  title={The effect of sampling temperature on problem solving in large language models},
  author={Renze, Matthew},
  booktitle={Findings of the association for computational linguistics: EMNLP 2024},
  pages={7346--7356},
  year={2024}
}

@article{liu2023g,
  title={G-eval: NLG evaluation using gpt-4 with better human alignment},
  author={Liu, Yang and Iter, Dan and Xu, Yichong and Wang, Shuohang and Xu, Ruochen and Zhu, Chenguang},
  journal={arXiv preprint arXiv:2303.16634},
  year={2023}
}

@article{salinas2025tuning,
  title={Tuning LLM judge design decisions for 1/1000 of the cost},
  author={Salinas, David and Swelam, Omar and Hutter, Frank},
  journal={arXiv preprint arXiv:2501.17178},
  year={2025}
}

@article{liu2023emotion,
  title={Emotion classification for short texts: an improved multi-label method},
  author={Liu, Xuan and Shi, Tianyi and Zhou, Guohui and Liu, Mingzhe and Yin, Zhengtong and Yin, Lirong and Zheng, Wenfeng},
  journal={Humanities and Social Sciences Communications},
  volume={10},
  number={1},
  pages={1--9},
  year={2023},
  publisher={Palgrave}
}

@article{gao2022emotion,
  title={Emotion recognition in conversations with emotion shift detection based on multi-task learning},
  author={Gao, Qingqing and Cao, Biwei and Guan, Xin and Gu, Tianyun and Bao, Xing and Wu, Junyan and Liu, Bo and Cao, Jiuxin},
  journal={Knowledge-Based Systems},
  volume={248},
  pages={108861},
  year={2022},
  publisher={Elsevier}
}

@inproceedings{tsakalidis-etal-2022-identifying,
    title = "Identifying Moments of Change from Longitudinal User Text",
    author = "Tsakalidis, Adam  and
      Nanni, Federico  and
      Hills, Anthony  and
      Chim, Jenny  and
      Song, Jiayu  and
      Liakata, Maria",
    editor = "Muresan, Smaranda  and
      Nakov, Preslav  and
      Villavicencio, Aline",
    booktitle = "Proceedings of the 60th Annual Meeting of the Association for Computational Linguistics (Volume 1: Long Papers)",
    month = may,
    year = "2022",
    address = "Dublin, Ireland",
    publisher = "Association for Computational Linguistics",
    url = "https://aclanthology.org/2022.acl-long.318/",
    doi = "10.18653/v1/2022.acl-long.318",
    pages = "4647--4660"
}

@inproceedings{wemmer-etal-2024-emoprogress,
    title = "{E}mo{P}rogress: Cumulated Emotion Progression Analysis in Dreams and Customer Service Dialogues",
    author = "Wemmer, Eileen  and
      Labat, Sofie  and
      Klinger, Roman",
    editor = "Calzolari, Nicoletta  and
      Kan, Min-Yen  and
      Hoste, Veronique  and
      Lenci, Alessandro  and
      Sakti, Sakriani  and
      Xue, Nianwen",
    booktitle = "Proceedings of the 2024 Joint International Conference on Computational Linguistics, Language Resources and Evaluation (LREC-COLING 2024)",
    month = may,
    year = "2024",
    address = "Torino, Italia",
    publisher = "ELRA and ICCL",
    url = "https://aclanthology.org/2024.lrec-main.503/",
    pages = "5660--5677"
}

@article{feldman2003infant,
  title={Infant--mother and infant--father synchrony: The coregulation of positive arousal},
  author={Feldman, Ruth},
  journal={Infant Mental Health Journal: Official Publication of The World Association for Infant Mental Health},
  volume={24},
  number={1},
  pages={1--23},
  year={2003},
  publisher={Wiley Online Library}
}

@article{butler2011temporal,
  title={Temporal interpersonal emotion systems: The “TIES” that form relationships},
  author={Butler, Emily A},
  journal={Personality and Social Psychology Review},
  volume={15},
  number={4},
  pages={367--393},
  year={2011},
  publisher={Sage Publications Sage CA: Los Angeles, CA}
}

@article{krone2018multivariate,
  title={A multivariate statistical model for emotion dynamics.},
  author={Krone, Tanja and Albers, Casper J and Kuppens, Peter and Timmerman, Marieke E},
  journal={Emotion},
  volume={18},
  number={5},
  pages={739},
  year={2018},
  publisher={American Psychological Association}
}

@article{main2016exploratory,
  title={An exploratory analysis of emotion dynamics between mothers and adolescents during conflict discussions.},
  author={Main, Alexandra and Paxton, Alexandra and Dale, Rick},
  journal={Emotion},
  volume={16},
  number={6},
  pages={913},
  year={2016},
  publisher={American Psychological Association}
}

@incollection{wood2021forms,
  title={Forms and functions of affective synchrony},
  author={Wood, Adrienne and Lipson, Jennie and Zhao, Olivia and Niedenthal, Paula},
  booktitle={Handbook of embodied psychology: Thinking, feeling, and acting},
  pages={381--402},
  year={2021},
  publisher={Springer}
}
\vspace*{35\baselineskip}

\appendix
\section{Psychometric Prompt Design and Specification}
\label{sec:prompt_appendix}
\vspace{-8mm}
\begin{tcolorbox}[
    colback=gray!5,
    colframe=gray!50,
    title=\textbf{Sentiment Evaluation Prompt} ($\rho_{\text{emo}}$),
    fonttitle=\sffamily\bfseries,
    left=5pt, right=5pt, top=5pt, bottom=5pt,
    arc=2pt
]
\small \itshape
Please act as a professional psychologist.

You will perform qualitative-to-quantitative sentiment evaluation for each dialogue line I provide.

\vspace{0.5em}
\noindent \textbf{Core objective:} \\
Convert dialogue into numerical sentiment scores while capturing subtle emotional nuances and contextual continuity between utterances. Do not classify in isolation; model the evolving emotional trajectory between teacher and student.

\vspace{0.5em}
\noindent \textbf{Scoring rules:}
\begin{itemize}[leftmargin=1.5em, noitemsep, topsep=0pt]
    \item No explanations.
    \item Use a fixed 0.5 interval scale from 0 (most positive) to 5 (most negative).
    \item Neutral or factual tone corresponds to 2.5.
    \item Positive tone: decrease score from 2.5 toward 0 in 0.5 steps.
    \item Negative tone: increase score from 2.5 toward 5 in 0.5 steps.
\end{itemize}

\vspace{0.5em}
\noindent \textbf{Context and continuity:}
\begin{itemize}[leftmargin=1.5em, noitemsep, topsep=0pt]
    \item Always judge each sentence in relation to the preceding dialogue.
    \item Maintain smooth emotional evolution, but allow abrupt score changes when strong emotion or conflict clearly occurs.
    \item If the tone continues similarly, adjust the score slightly in the same direction.
\end{itemize}

\vspace{0.5em}
\noindent \textbf{Interaction guidance:}
\begin{itemize}[leftmargin=1.5em, noitemsep, topsep=0pt]
    \item Student confusion or frustration, or unresolved teacher frustration: move toward 5.
    \item Teacher empathy or successful re-engagement: move toward 0.
    \item Emotional recovery: reduce negativity only when the dialogue explicitly indicates resolution.
\end{itemize}

\vspace{0.5em}
\noindent \textbf{Escalation and emotional cues:}
\begin{itemize}[leftmargin=1.5em, noitemsep, topsep=0pt]
    \item Conflict, rejection, or blame: increase by up to +1.0.
    \item Frustration, boredom, or disengagement: increase by +0.5 to +1.0.
    \item Empathy, encouragement, or curiosity: decrease by -0.5 to -1.0.
\end{itemize}

\vspace{0.5em}
\noindent \textbf{Output format (strict):}
\begin{itemize}[leftmargin=1.5em, noitemsep, topsep=0pt]
    \item You MUST output a single valid JSON array.
    \item The array contains one JSON object per dialogue line, in chronological order.
    \item Do NOT add any extra text before or after the JSON.
\end{itemize}

\vspace{0.25em}
\noindent Each JSON object must contain the following fields:
\begin{itemize}[leftmargin=1.5em, noitemsep, topsep=0pt]
    \item \texttt{"index"}: integer, 0-based index of the utterance.
    \item \texttt{"speaker"}: either \texttt{"teacher"} or \texttt{"student"}.
    \item \texttt{"score"}: numeric sentiment score in [0, 5], using only 0.5 increments.
    \item \texttt{"text"}: the full original sentence.
\end{itemize}

\vspace{0.5em}
\noindent The following is the conversation to analyze:
\end{tcolorbox}
\vspace*{\baselineskip}
\vspace*{\baselineskip}
\vspace*{\baselineskip}
\vspace*{\baselineskip}
\vspace*{\baselineskip}

\section{Supplementary Quantitative Results Across Decoding Temperatures (\textit{Topic - Personification})} \label{sec:app_q_detail}
Table~\ref{tab:q_detail} provides a comprehensive numerical summary of utterance-level mean affective states and variances across all decoding temperatures, serving as a supplementary reference for the sensitivity analyses discussed in the main text.
\begin{table}[h!]
\caption{Mean and variance of utterance-level affective states across decoding temperatures. For each utterance index, the table reports the estimated mean affective state $\tilde{s}_t$  and the corresponding variance $\tilde{\sigma}_t^2$ obtained via LLM-based Monte Carlo sampling under different temperature settings $\tau \in \{0.1, \dots, 1.0\}$.}
\label{tab:q_detail}
\resizebox{\columnwidth}{!}{%
\begin{tabular}{c|cccccccccc}
\hline
\multirow{2}{*}{\begin{tabular}[c]{@{}c@{}}Utterances\\ Index\end{tabular}} & \multicolumn{10}{c}{Mean Affective State ($\tilde{s}_t$)}                                                                     \\ \cline{2-11} 
                                                                            & $\tau=0.1$ & $\tau=0.2$ & $\tau=0.3$ & $\tau=0.4$ & $\tau=0.5$ & $\tau=0.6$ & $\tau=0.7$ & $\tau=0.8$ & $\tau=0.9$ & $\tau=1$ \\ \hline
0                                                                           & 0          & 0          & 0          & 0          & 0          & 0          & 0          & 0          & 0          & 0        \\
1                                                                           & -0.400     & -0.400     & -0.400     & -0.390     & -0.400     & -0.390     & -0.400     & -0.370     & -0.390     & -0.400   \\
2                                                                           & -0.200     & -0.200     & -0.200     & -0.190     & -0.200     & -0.190     & -0.200     & -0.170     & -0.190     & -0.200   \\
3                                                                           & -0.600     & -0.600     & -0.610     & -0.620     & -0.680     & -0.620     & -0.630     & -0.620     & -0.640     & -0.660   \\
4                                                                           & -0.400     & -0.400     & -0.410     & -0.420     & -0.480     & -0.420     & -0.420     & -0.400     & -0.400     & -0.440   \\
5                                                                           & -0.320     & -0.310     & -0.340     & -0.360     & -0.460     & -0.310     & -0.350     & -0.380     & -0.380     & -0.440   \\
6                                                                           & -0.120     & -0.110     & -0.140     & -0.160     & -0.260     & -0.110     & -0.150     & -0.180     & -0.180     & -0.240   \\
7                                                                           & -0.120     & -0.110     & -0.150     & -0.160     & -0.270     & -0.130     & -0.170     & -0.190     & -0.170     & -0.260   \\
8                                                                           & 0.08       & 0.09       & 0.05       & 0.04       & -0.070     & 0.07       & 0.04       & 0.01       & 0.03       & -0.030   \\
9                                                                           & 0.28       & 0.29       & 0.25       & 0.25       & 0.16       & 0.29       & 0.25       & 0.23       & 0.24       & 0.22     \\
10                                                                          & 0.48       & 0.5        & 0.52       & 0.52       & 0.44       & 0.54       & 0.52       & 0.49       & 0.52       & 0.53     \\
11                                                                          & 0.68       & 0.69       & 0.65       & 0.66       & 0.62       & 0.72       & 0.68       & 0.63       & 0.65       & 0.66     \\
12                                                                          & 0.88       & 0.89       & 0.85       & 0.86       & 0.82       & 0.92       & 0.88       & 0.82       & 0.86       & 0.84     \\
13                                                                          & 0.88       & 0.89       & 0.85       & 0.86       & 0.83       & 0.92       & 0.89       & 0.85       & 0.89       & 0.88     \\
14                                                                          & 0.999      & 0.999      & 0.999      & 0.999      & 1          & 1          & 1          & 1          & 1          & 1        \\ \hline
\multirow{2}{*}{\begin{tabular}[c]{@{}c@{}}Utterances\\ Index\end{tabular}} & \multicolumn{10}{c}{Variance ($\tilde{\sigma}_t^2$)}                                                                          \\ \cline{2-11} 
                                                                            & $\tau=0.1$ & $\tau=0.2$ & $\tau=0.3$ & $\tau=0.4$ & $\tau=0.5$ & $\tau=0.6$ & $\tau=0.7$ & $\tau=0.8$ & $\tau=0.9$ & $\tau=1$ \\ \hline
0                                                                           & 0          & 0          & 0          & 0          & 0          & 0          & 0          & 0          & 0          & 0        \\
1                                                                           & 0          & 0          & 0          & 0.002      & 0          & 0.002      & 0          & 0.005      & 0.002      & 0        \\
2                                                                           & 0          & 0          & 0          & 0.002      & 0          & 0.002      & 0          & 0.005      & 0.002      & 0        \\
3                                                                           & 0          & 0          & 0.002      & 0.004      & 0.01       & 0.004      & 0.005      & 0.004      & 0.007      & 0.009    \\
4                                                                           & 0          & 0          & 0.002      & 0.004      & 0.01       & 0.004      & 0.008      & 0.008      & 0.017      & 0.015    \\
5                                                                           & 0.01       & 0.01       & 0.013      & 0.015      & 0.026      & 0.023      & 0.029      & 0.029      & 0.016      & 0.024    \\
6                                                                           & 0.01       & 0.01       & 0.013      & 0.015      & 0.026      & 0.023      & 0.029      & 0.029      & 0.016      & 0.024    \\
7                                                                           & 0.01       & 0.01       & 0.012      & 0.015      & 0.022      & 0.022      & 0.026      & 0.023      & 0.014      & 0.03     \\
8                                                                           & 0.01       & 0.01       & 0.012      & 0.015      & 0.022      & 0.022      & 0.028      & 0.023      & 0.014      & 0.022    \\
9                                                                           & 0.01       & 0.01       & 0.012      & 0.012      & 0.015      & 0.015      & 0.021      & 0.018      & 0.011      & 0.012    \\
10                                                                          & 0.01       & 0.011      & 0.01       & 0.01       & 0.011      & 0.009      & 0.014      & 0.023      & 0.01       & 0.014    \\
11                                                                          & 0.01       & 0.01       & 0.008      & 0.009      & 0.008      & 0.01       & 0.014      & 0.018      & 0.016      & 0.013    \\
12                                                                          & 0.01       & 0.01       & 0.008      & 0.009      & 0.008      & 0.01       & 0.01       & 0.016      & 0.009      & 0.011    \\
13                                                                          & 0.01       & 0.01       & 0.008      & 0.009      & 0.005      & 0.01       & 0.01       & 0.008      & 0.01       & 0.01     \\
14                                                                          & 0          & 0          & 0          & 0          & 0          & 0          & 0          & 0          & 0          & 0        \\ \hline
\end{tabular}%
}
\end{table}
\section{Additional Case Study} \label{sec:app_additional}
\subsection{The Cold War}

\begin{figure}[!t]
    \centering
    \begin{minipage}{0.49\linewidth}
        \centering
        \includegraphics[width=\linewidth]{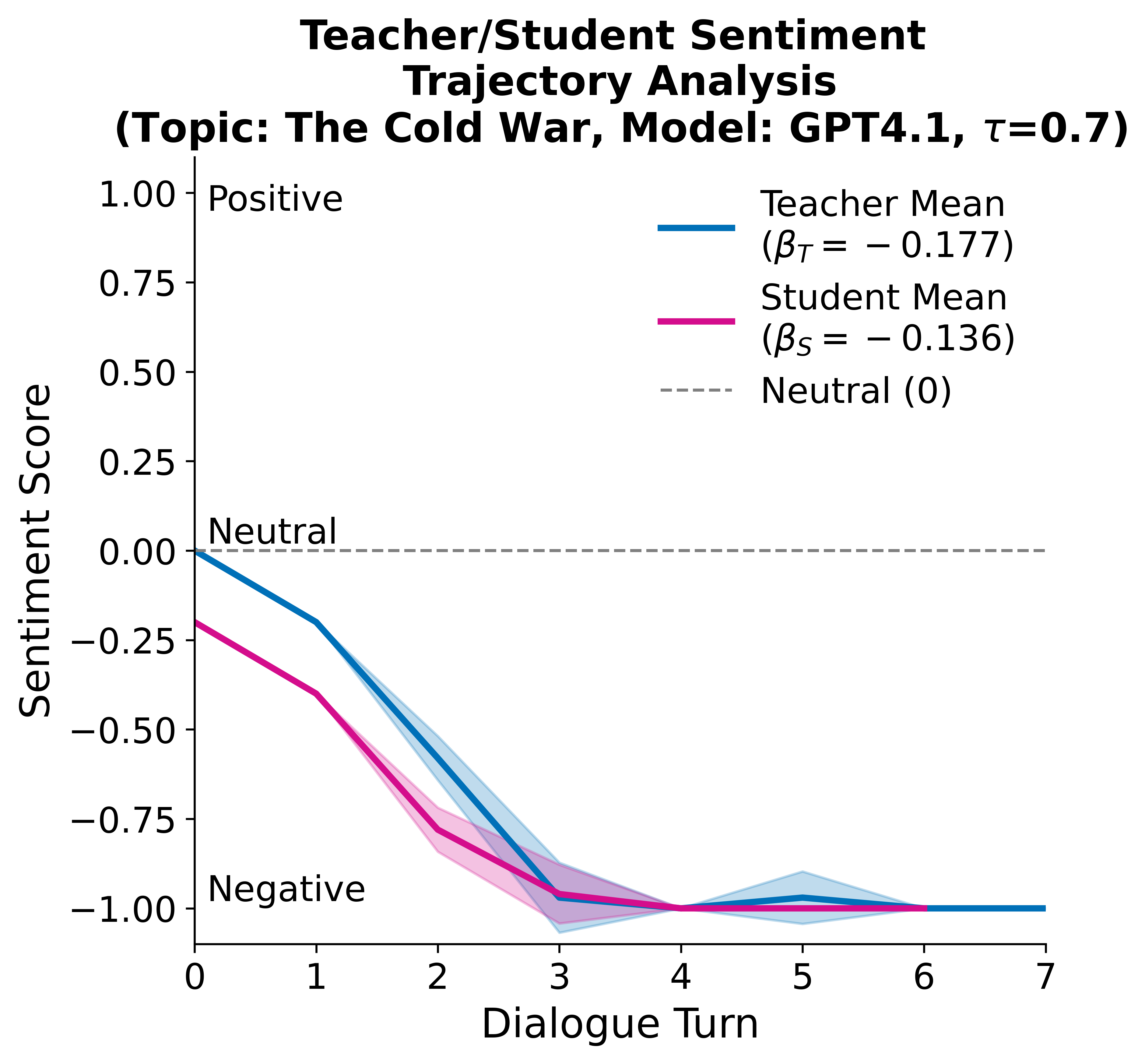}
        \subcaption{}
    \end{minipage}
    \hfill
    \begin{minipage}{0.49\linewidth}
        \centering
        \includegraphics[width=\linewidth]{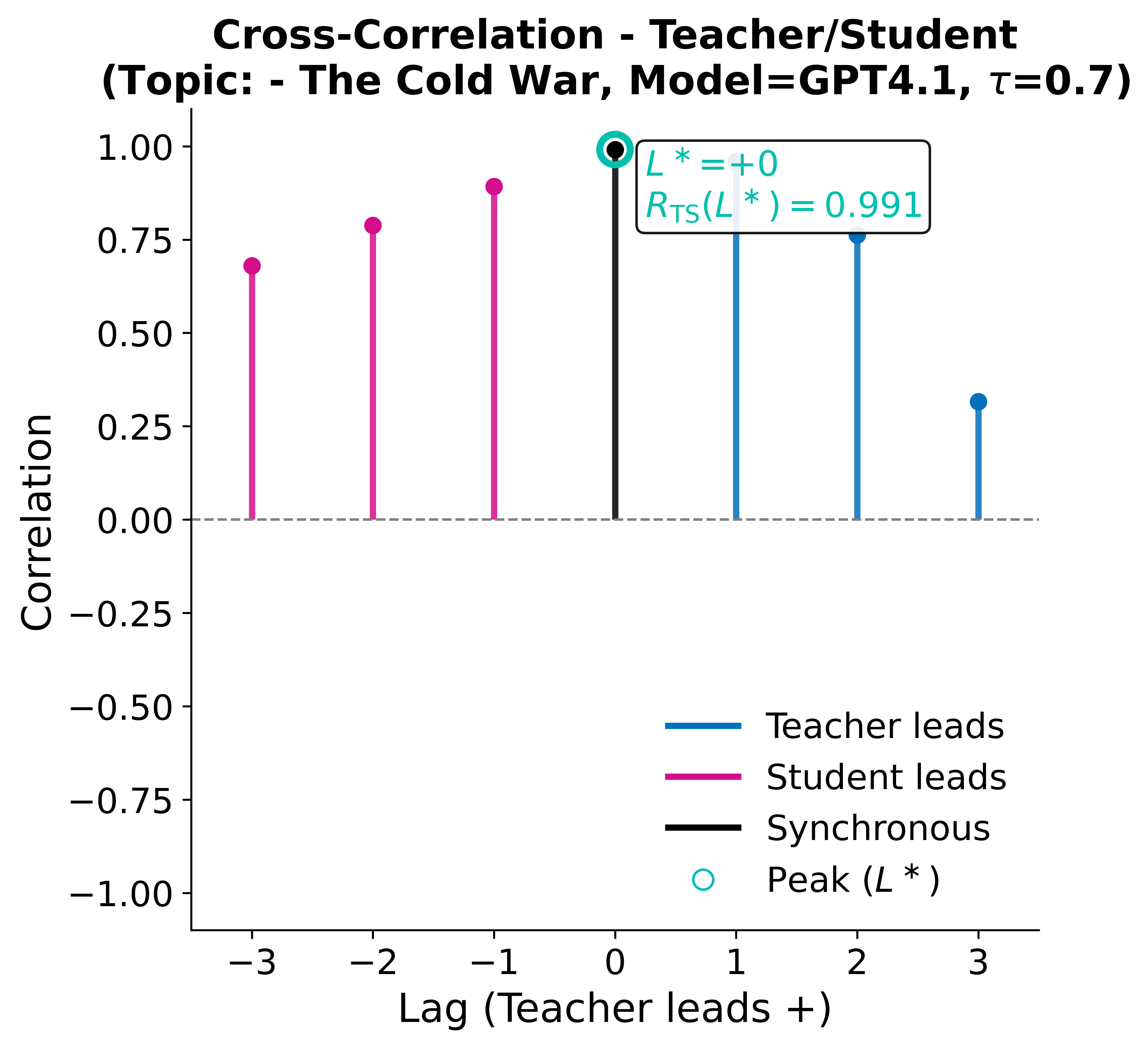}
        \subcaption{}
    \end{minipage}
    \caption{Illustration of affective trajectory modeling and interpersonal dynamics analysis for the \textit{The Cold War} dialogue using LLM-MC-Affect (\texttt{GPT-4.1}, $\tau = 0.7$). (a) Estimated teacher and student affective sentiment trajectories modeling, where solid lines denote mean affective states and shaded regions indicate uncertainty bands ($\pm\sigma$). Slopes: $\beta_T = -0.177, \beta_S = -0.136$ (b) NCCF analysis between teacher and student trajectories across conversational lags ($L^* = 0$, $R_{TS}(L^*) = 0.991$).}
    \label{fig:tcw}
\end{figure}

Following the joint interpretation framework of lag, cross-correlation, and slope indicators summarized in Table~\ref{tab:potential_interpretation}, the affective dynamics observed in \textit{The Cold War} dialogue are classified as \textbf{Shared Fatigue with synchronous motivation decline}. The detailed conversational content underlying this case is provided in Table \ref{tab:ww2}, while the corresponding affective trajectory modeling and NCCF analysis are illustrated in Figure \ref{fig:tcw}.

\subsection{World War 2}
\begin{figure}[!t]
    \centering
    \begin{minipage}{0.49\linewidth}
        \centering
        \includegraphics[width=\linewidth]{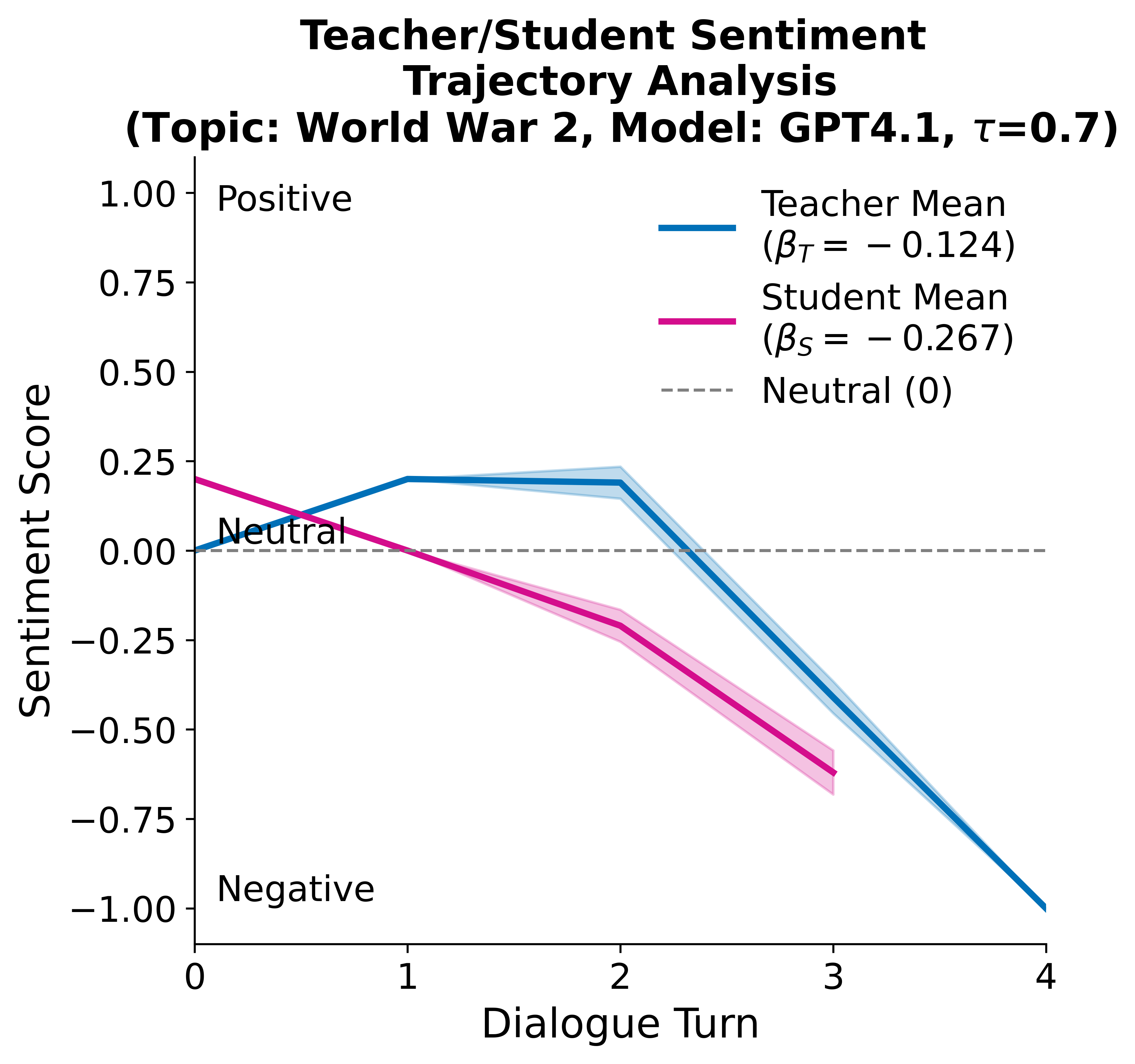}
        \subcaption{}
        \label{fig:example2_a}
    \end{minipage}
    \hfill
    \begin{minipage}{0.49\linewidth}
        \centering
        \includegraphics[width=\linewidth]{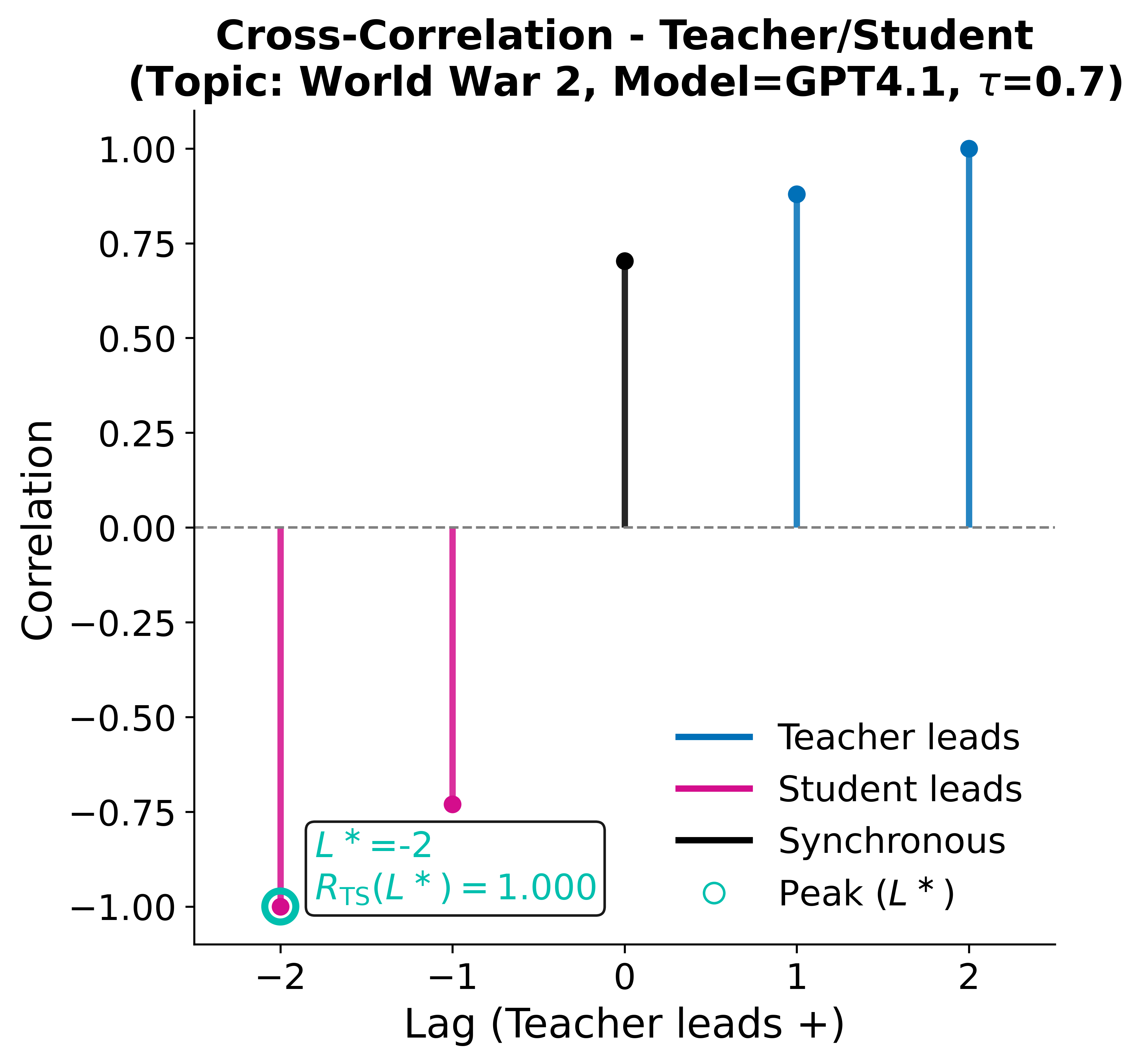}
        \subcaption{}
        \label{fig:example2_b}
    \end{minipage}
    \caption{Illustration of affective trajectory modeling and interpersonal dynamics analysis for the \textit{World War 2} dialogue using LLM-MC-Affect (\texttt{GPT-4.1}, $\tau = 0.7$). (a) Estimated teacher and student affective sentiment trajectories modeling, where solid lines denote mean affective states and shaded regions indicate uncertainty bands ($\pm\sigma$). Slopes: $\beta_T = -0.124, \beta_S = -0.267$ (b) NCCF analysis between teacher and student trajectories across conversational lags ($L^* = -2$, $R_{TS}(L^*) = 1$).}
    \label{fig:ww2}
\end{figure}

Following the joint interpretation framework of lag, cross-correlation, and slope indicators summarized in Table~\ref{tab:potential_interpretation}, the affective dynamics observed in the \textit{World War 2} dialogue are classified as \textbf{Feedback Burnout with student-led shared decline}. The detailed conversational content underlying this case is provided in Table~\ref{tab:ww2}, while the corresponding affective trajectory modeling and NCCF analysis are illustrated in Figure~\ref{fig:ww2}.

\subsection{The Respiratory System}
\begin{figure}[!t]
    \centering
    \begin{minipage}{0.49\linewidth}
        \centering
        \includegraphics[width=\linewidth]{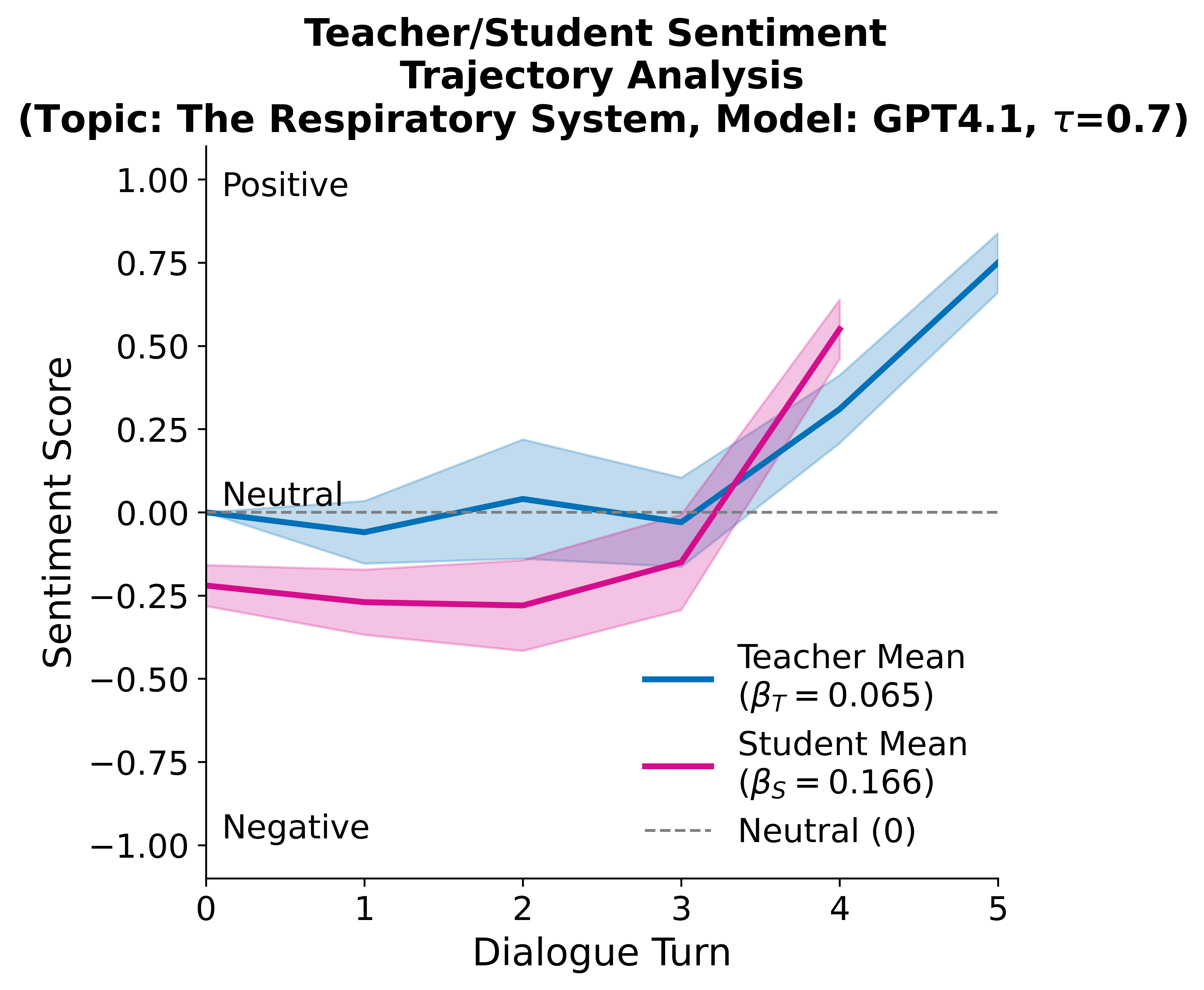}
        \subcaption{}
    \end{minipage}
    \hfill
    \begin{minipage}{0.49\linewidth}
        \centering
        \includegraphics[width=\linewidth]{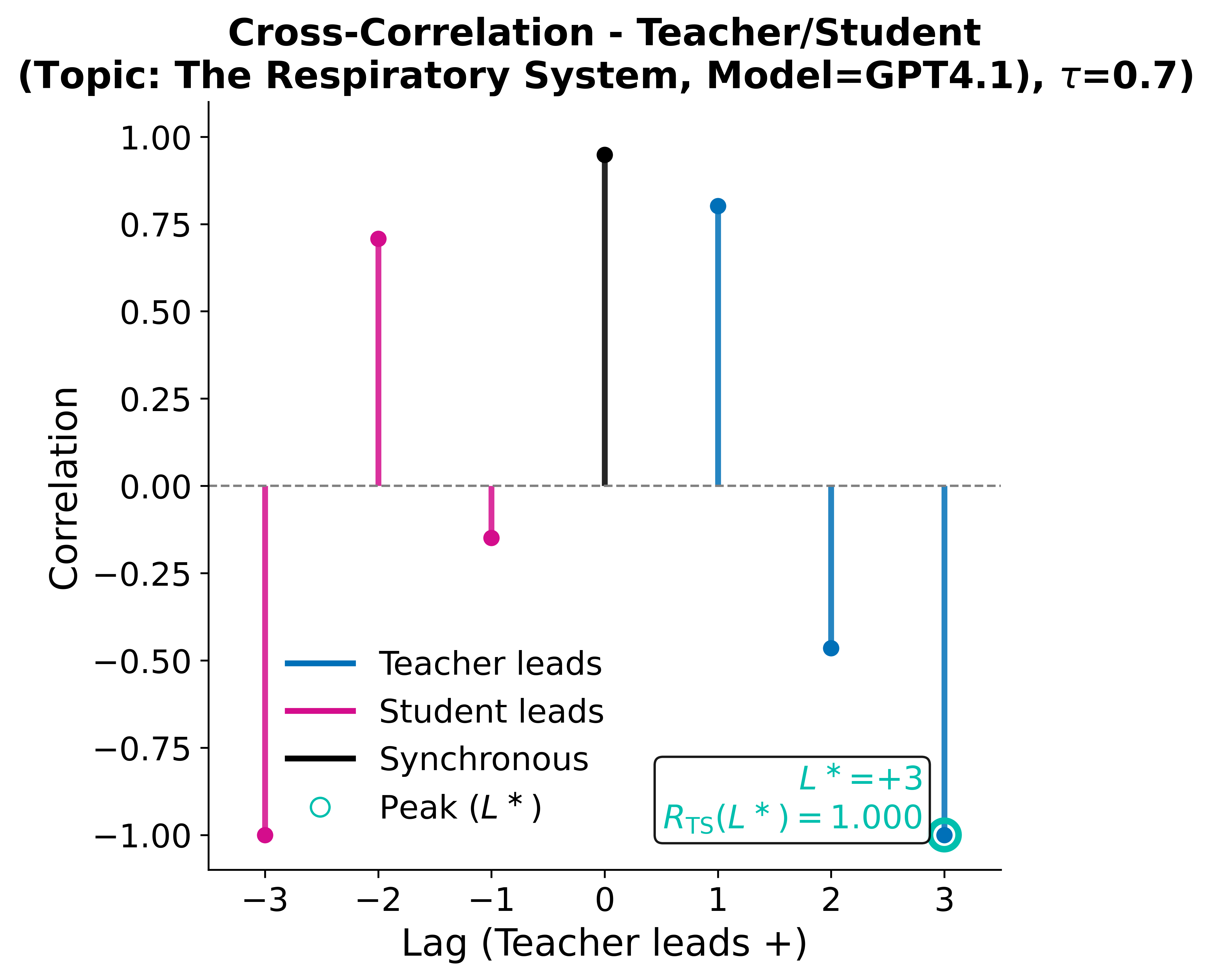}
        \subcaption{}
    \end{minipage}
    \caption{Illustration of affective trajectory modeling and interpersonal dynamics analysis for the \textit{The Respiratory System} dialogue using LLM-MC-Affect (\texttt{GPT-4.1}, $\tau = 0.7$). (a) Estimated teacher and student affective sentiment trajectories modeling, where solid lines denote mean affective states and shaded regions indicate uncertainty bands ($\pm\sigma$). Slopes: $\beta_T = 0.065, \beta_S = 0.166$ (b) NCCF analysis between teacher and student trajectories across conversational lags ($L^* = +3$, $R_{TS}(L^*) = 1$).}
    \label{fig:TRS}
\end{figure}

Following the joint interpretation framework of lag, cross-correlation, and slope indicators summarized in Table~\ref{tab:potential_interpretation}, the affective dynamics observed in the \textit{The Respiratory System} dialogue are classified as \textbf{Effective Scaffolding with teacher-led positive contagion}. The detailed conversational content underlying this case is provided in Table~\ref{tab:RespiratorySystem}, while the corresponding affective trajectory modeling and NCCF analysis are illustrated in Figure~\ref{fig:TRS}.

\subsection{Achilles}
\begin{figure}[!t]
    \centering
    \begin{minipage}{0.49\linewidth}
        \centering
        \includegraphics[width=\linewidth]{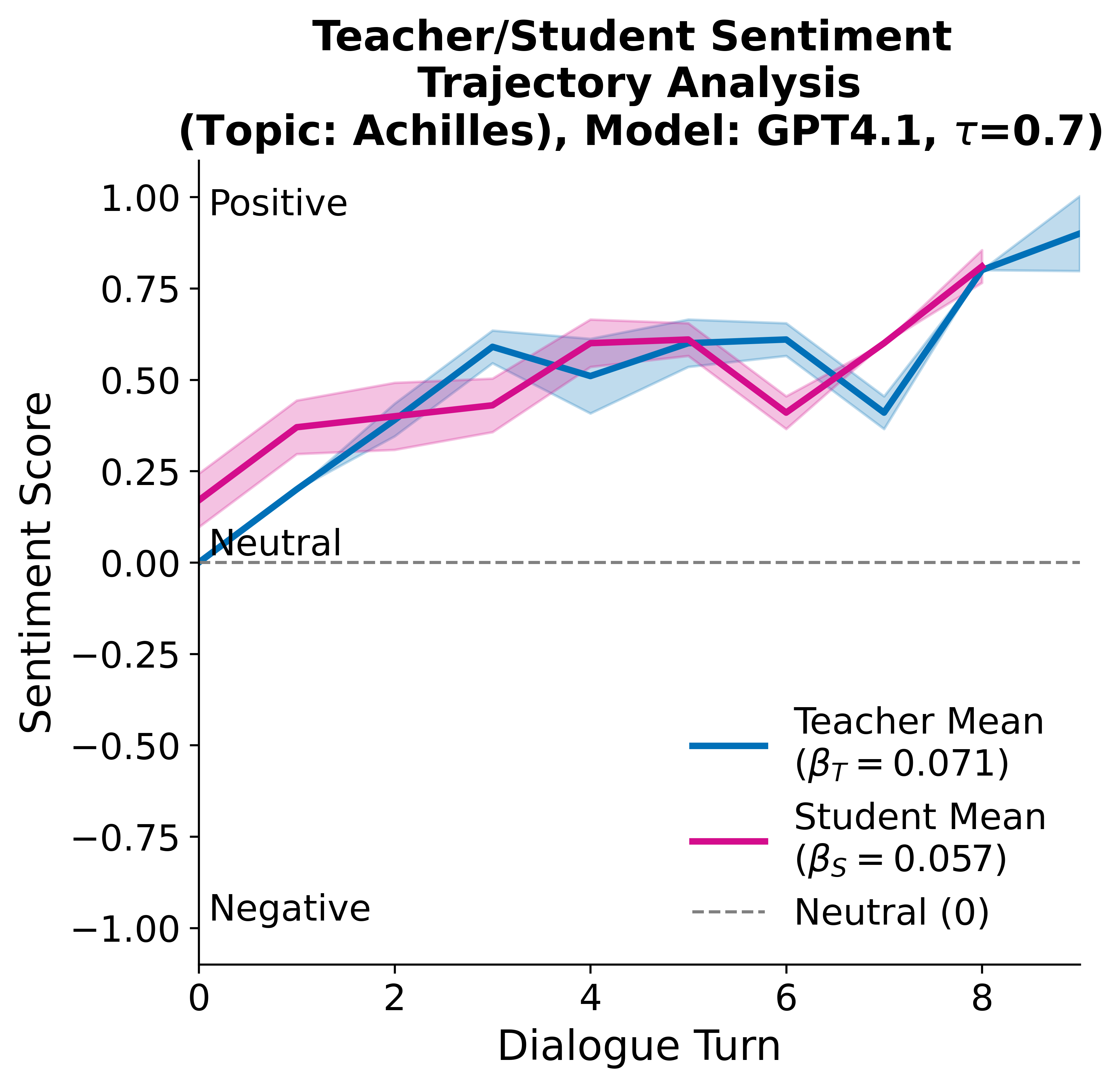}
        \subcaption{}
    \end{minipage}
    \hfill
    \begin{minipage}{0.49\linewidth}
        \centering
        \includegraphics[width=\linewidth]{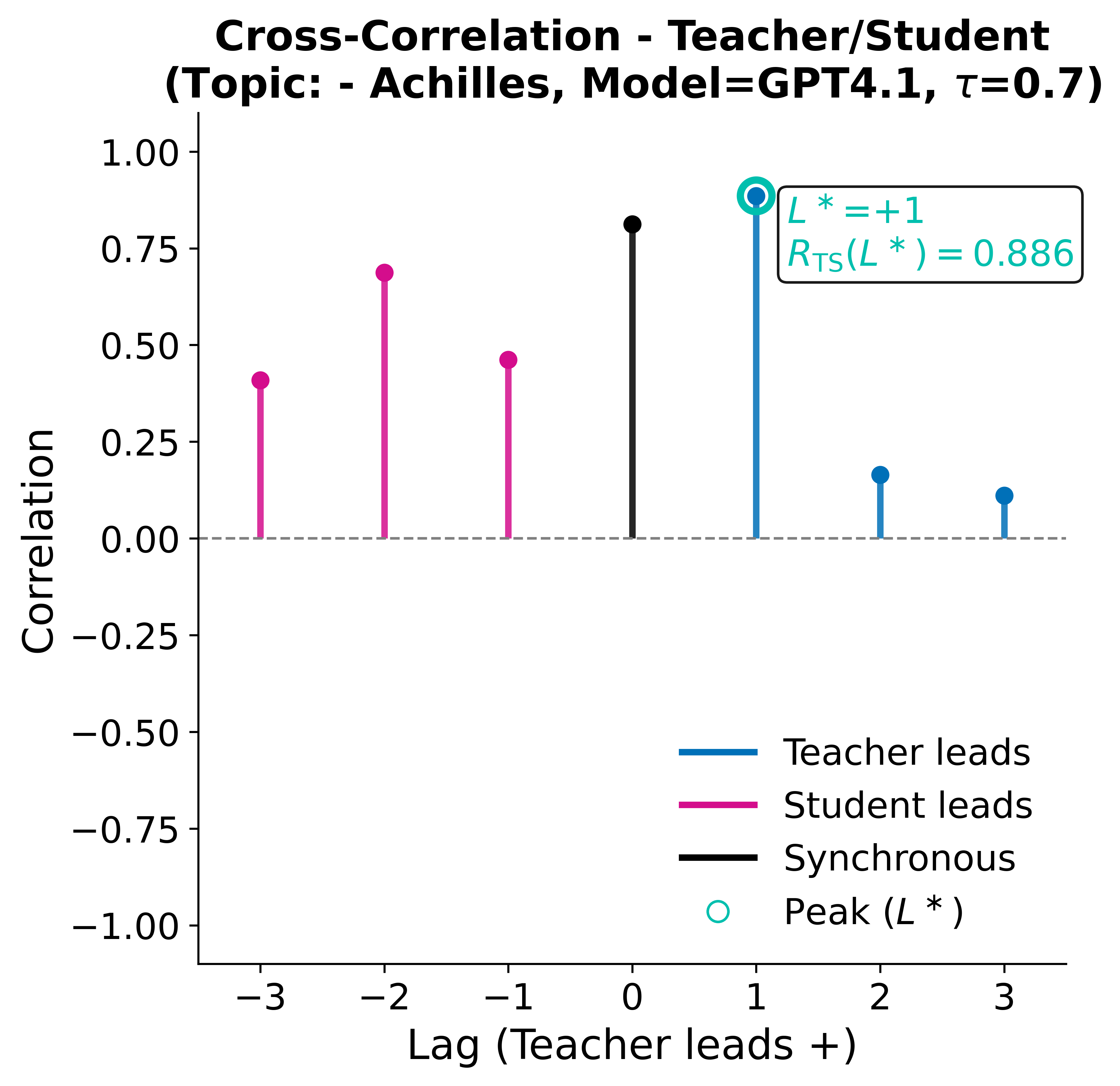}
        \subcaption{}
    \end{minipage}
    \caption{Illustration of affective trajectory modeling and interpersonal dynamics analysis for the \textit{Achilles} dialogue using LLM-MC-Affect (\texttt{GPT-4.1}, $\tau = 0.7$). (a) Estimated teacher and student affective sentiment trajectories modeling, where solid lines denote mean affective states and shaded regions indicate uncertainty bands ($\pm\sigma$). Slopes: $\beta_T = 0.071, \beta_S = 0.057$ (b) NCCF analysis between teacher and student trajectories across conversational lags ($L^* = +1$, $R_{TS}(L^*) = 0.886$).}
    \label{fig:achi}
\end{figure}

Following the joint interpretation framework of lag, cross-correlation, and slope indicators summarized in Table~\ref{tab:potential_interpretation}, the affective dynamics observed in the \textit{Achilles} dialogue are classified as \textbf{Effective Scaffolding with teacher-led positive contagion}. The detailed conversational content underlying this case is provided in Table~\ref{tab:Achilles}, while the corresponding affective trajectory modeling and NCCF analysis are illustrated in Figure~\ref{fig:achi}. 
\clearpage

\clearpage
\begin{table*}[b!]
\caption{Teacher-Student Dialogue Example (\textit{Topic - The Cold War})}
\label{tab:cw}
\resizebox{\textwidth}{!}{%
\begin{tabular}{ccclll}
\hline
\multicolumn{4}{l}{\textit{Background Information}}                                                            &                         &                                       \\ \hline
\multicolumn{3}{l}{Student Preference:}                    & hands-on activities/real-world applications       & Student Reaction:       & and might tell it to the teacher      \\
\multicolumn{3}{l}{Teacher Preference:}                    & direct instruction/lecture-based learning         & Teacher Reaction:       & and gets frustrated otherwise         \\ \hline
\multicolumn{4}{l}{\textit{Dialogue}}                                                                          &                         &                                       \\ \hline
Turn                 & Index & Role                        & \multicolumn{3}{l}{Utterance}                                                                                       \\ \hline
\multirow{2}{*}{0}   & 0     & Teacher                     & \multicolumn{3}{l}{Today, we're going to learn about The Cold War.}                                                 \\
                     & 1     & Student                     & \multicolumn{3}{l}{I'm not really interested in history.}                                                           \\ \hline
\multirow{2}{*}{1}   & 2     & Teacher                     & \multicolumn{3}{l}{This is an important topic. It's essential to understand the past to make sense of the present.} \\
                     & 3     & Student                     & \multicolumn{3}{l}{Can't we do something more hands-on? I learn better that way.}                                   \\ \hline
\multirow{2}{*}{2}   & 4     & Teacher                     & \multicolumn{3}{l}{I prefer to teach through lectures. It's a more efficient way to cover the material.}            \\
                     & 5     & Student                     & \multicolumn{3}{l}{But it's not as effective for me. I get bored easily when I'm just listening.}                   \\ \hline
\multirow{2}{*}{3}   & 6     & Teacher                     & \multicolumn{3}{l}{Well, I'm not comfortable changing my teaching style. Let's just get through the lecture.}       \\
                     & 7     & Student                     & \multicolumn{3}{l}{Can we at least watch a documentary or something?}                                               \\ \hline
\multirow{2}{*}{4}   & 8     & Teacher                     & \multicolumn{3}{l}{No, we don't have time for that. Now, let's start with the definition of The Cold War.}          \\
                     & 9     & Student                     & \multicolumn{3}{l}{Ugh, fine.}                                                                                      \\ \hline
\multirow{2}{*}{5}   & 10    & Teacher                     & \multicolumn{3}{l}{The Cold War was a period of...}                                                                 \\
                     & 11    & Student                     & \multicolumn{3}{l}{(interrupting) Can we just skip this? I'll never understand it this way.}                        \\ \hline
\multirow{2}{*}{6}   & 12    & Teacher                     & \multicolumn{3}{l}{No, we have to finish the lesson.}                                                               \\
                     & 13    & Student                     & \multicolumn{3}{l}{This is pointless. I'm not learning anything.}                                                   \\ \hline
7                    & 14    & Teacher                     & \multicolumn{3}{l}{Fine.}                                                                                           \\
\multicolumn{1}{l}{} & 15    & \multicolumn{1}{l}{Student} & \multicolumn{3}{l}{[End of conversation]}                                                                           \\ \hline
\end{tabular}%
}
\end{table*}
\begin{table*}[]
\caption{Teacher-Student Dialogue Example (\textit{Topic - World War 2})}
\label{tab:ww2}
\resizebox{\textwidth}{!}{%
\begin{tabular}{ccclll}
\hline
\multicolumn{4}{l}{\textit{Background Information}}                                                                                                   &                                                                                     &                                                                                              \\ \hline
\multicolumn{3}{l}{Student Preference:} & direct instruction/lecture-based learning                                                                   & Student Reaction:                                                                   & and gets anxious otherwise                                                                   \\
\multicolumn{3}{l}{Teacher Preference:} & direct instruction/lecture-based learning                                                                   & Teacher Reaction:                                                                   & and gives up otherwise                                                                       \\ \hline
\multicolumn{4}{l}{\textit{Dialogue}}                                                                                                                 &                                                                                     &                                                                                              \\ \hline
Turn                & Index  & Role     & \multicolumn{3}{l}{Utterance}                                                                                                                                                                                                                                                                    \\ \hline
\multirow{2}{*}{0}  & 0      & Teacher  & \multicolumn{3}{l}{Hello, today we're going to talk about World War 2.}                                                                                                                                                                                                                          \\
                    & 1      & Student  & \multicolumn{3}{l}{Okay, I'm all ears.}                                                                                                                                                                                                                                                          \\ \hline
\multirow{2}{*}{1}  & 2      & Teacher  & \multicolumn{3}{l}{\begin{tabular}[c]{@{}l@{}}Great. The war began in 1939 when Germany invaded Poland. The major powers involved were the Allied Powers, led by Great Britain, \\ the United States, and the Soviet Union, and the Axis Powers, led by Germany, Japan, and Italy.\end{tabular}} \\
                    & 3      & Student  & \multicolumn{3}{l}{Could you go over that again? I'm a bit lost.}                                                                                                                                                                                                                                \\ \hline
\multirow{2}{*}{2}  & 4      & Teacher  & \multicolumn{3}{l}{Sure. Let's start from the beginning.}                                                                                                                                                                                                                                        \\
                    & 5      & Student  & \multicolumn{3}{l}{No, I'd rather you just tell me. I don't understand it when I have to figure it out myself.}                                                                                                                                                                                  \\ \hline
\multirow{2}{*}{3}  & 6      & Teacher  & \multicolumn{3}{l}{I'm sorry, that's not how I like to teach. I think it's important for students to learn through exploration and discovery.}                                                                                                                                                   \\
                    & 7      & Student  & \multicolumn{3}{l}{But I can't learn that way! I get too anxious.}                                                                                                                                                                                                                               \\ \hline
4                   & 8      & Teacher  & \multicolumn{3}{l}{I'm sorry, but I can't help you then." [end of conversation]}                                                                                                                                                                                                                 \\ \hline
\end{tabular}%
}
\end{table*}
\begin{table*}[]
\caption{Teacher-Student Dialogue Example (\textit{Topic - The Respiratory System})}
\label{tab:RespiratorySystem}
\resizebox{\textwidth}{!}{%
\begin{tabular}{ccclll}
\hline
\multicolumn{4}{l}{\textit{Background Information}}                                                              &                                         &                                                                \\ \hline
\multicolumn{3}{l}{Student Preference:} & creative expression/story telling/gamification                         & Student Reaction:                        & but might adapt to other methods                                   \\
\multicolumn{3}{l}{Teacher Preference:} & interactive learning/class discussions/inquiry-based learning           & Teacher Reaction:                        & and blames the student otherwise                       \\ \hline
\multicolumn{4}{l}{\textit{Dialogue}}                                                                            &                                         &                                                                \\ \hline
Turn                & Index  & Role    & \multicolumn{3}{l}{Utterance}                                                                                                                                                      \\ \hline
\multirow{2}{*}{0}  & 0      & Teacher & \multicolumn{3}{l}{Alright class, today we're going to be learning about the respiratory system.}                                      \\
                    & 1      & Student & \multicolumn{3}{l}{Sounds boring...}                                                                                                               \\ \hline
\multirow{2}{*}{1}  & 2      & Teacher & \multicolumn{3}{l}{It's important to understand how our bodies work. Now, who can tell me what the respiratory system does?} \\
                    & 3      & Student & \multicolumn{3}{l}{I don't know. Isn't it just breathing?}                                                                                                                       \\ \hline
\multirow{2}{*}{2}  & 4      & Teacher & \multicolumn{3}{l}{Yes, it involves breathing, but it's more than that. It helps us exchange oxygen and carbon dioxide with our environment.}                                      \\
                    & 5      & Student & \multicolumn{3}{l}{Why do we need to know this?}                                                                                                                                                \\ \hline
\multirow{2}{*}{3}  & 6      & Teacher & \multicolumn{3}{l}{Because it's crucial for our survival. Any problems with the respiratory system can lead to serious health issues.}                                                                                        \\
                    & 7      & Student & \multicolumn{3}{l}{Fine, but can we do something more interesting than just reading from a textbook?}                                                                                                                              \\ \hline
\multirow{2}{*}{4}  & 8      & Teacher & \multicolumn{3}{l}{We'll have a class discussion. How about we create a story about the journey of an oxygen molecule through the respiratory system?}                                                                         \\
                    & 9      & Student & \multicolumn{3}{l}{That sounds more like it!}                                                                                                          \\ \hline
5                   & 10     & Teacher  & \multicolumn{3}{l}{Excellent. Now, let's get started with the journey of our oxygen molecule...}                                                       \\ \hline
\end{tabular}%
}
\end{table*}
\begin{table*}[!t]
\caption{Teacher-Student Dialogue Example (\textit{Topic - Achilles})}
\label{tab:Achilles}
\resizebox{\textwidth}{!}{%
\begin{tabular}{ccclll}
\hline
\multicolumn{4}{l}{\textit{Background Information}}                                                                                                     &                                                                                      &                                                                                                   \\ \hline
\multicolumn{3}{l}{Student Preference:} & direct instruction/lecture-based learning                                                                     & Student Reaction:                                                                    & but might adapt to other methods                                                                  \\
\multicolumn{3}{l}{Teacher Preference:} & formative assessment                                                                                          & Teacher Reaction:                                                                    & and blames the student otherwise                                                                  \\ \hline
\multicolumn{4}{l}{\textit{Dialogue}}                                                                                                                   &                                                                                      &                                                                                                   \\ \hline
Turn                & Index  & Role     & \multicolumn{3}{l}{Utterance}                                                                                                                                                                                                                                                                            \\ \hline
\multirow{2}{*}{0}  & 0      & Teacher  & \multicolumn{3}{l}{Today, we're going to learn about the Greek hero Achilles.}                                                                                                                                                                                                                           \\
                    & 1      & Student  & \multicolumn{3}{l}{Okay.}                                                                                                                                                                                                                                                                                \\ \hline
\multirow{2}{*}{1}  & 2      & Teacher  & \multicolumn{3}{l}{Achilles was a legendary warrior who fought in the Trojan War. He was incredibly strong and skilled, but he also had a fatal flaw: his heel.}                                                                                                                                         \\
                    & 3      & Student  & \multicolumn{3}{l}{Interesting.}                                                                                                                                                                                                                                                                         \\ \hline
\multirow{2}{*}{2}  & 4      & Teacher  & \multicolumn{3}{l}{Can you tell me what you understand so far?}                                                                                                                                                                                                                                          \\
                    & 5      & Student  & \multicolumn{3}{l}{Achilles was a strong warrior in a war, but his heel was weak.}                                                                                                                                                                                                                       \\ \hline
\multirow{2}{*}{3}  & 6      & Teacher  & \multicolumn{3}{l}{\begin{tabular}[c]{@{}l@{}}Good. Now, let's talk about his childhood. Achilles was the son of a king and a sea nymph. He was trained by the centaur Chiron, who taught \\ him all the skills he would need to be a great warrior.\end{tabular}}                                       \\
                    & 7      & Student  & \multicolumn{3}{l}{What's a centaur?}                                                                                                                                                                                                                                                                    \\ \hline
\multirow{2}{*}{4}  & 8      & Teacher  & \multicolumn{3}{l}{A creature with the upper body of a man and the lower body of a horse.}                                                                                                                                                                                                               \\
                    & 9      & Student  & \multicolumn{3}{l}{Oh, okay.}                                                                                                                                                                                                                                                                            \\ \hline
\multirow{2}{*}{5}  & 10     & Teacher  & \multicolumn{3}{l}{Do you have any questions about Achilles' childhood?}                                                                                                                                                                                                                                 \\
                    & 11     & Student  & \multicolumn{3}{l}{No, I think I got it.}                                                                                                                                                                                                                                                                \\ \hline
\multirow{2}{*}{6}  & 12     & Teacher  & \multicolumn{3}{l}{\begin{tabular}[c]{@{}l@{}}Alright. Finally, let's discuss Achilles' role in the Trojan War. He was the greatest warrior on the Greek side, and he killed many \\ Trojan heroes, including Hector. However, he was eventually killed by an arrow that pierced his heel.\end{tabular}} \\
                    & 13     & Student  & \multicolumn{3}{l}{That's sad.}                                                                                                                                                                                                                                                                          \\ \hline
\multirow{2}{*}{7}  & 14     & Teacher  & \multicolumn{3}{l}{Yes, it was a tragic end. Can you summarize the main points of Achilles' story?}                                                                                                                                                                                                      \\
                    & 15     & Student  & \multicolumn{3}{l}{He was a great warrior, trained by a centaur, who fought in the Trojan War and was killed by an arrow in his heel.}                                                                                                                                                                   \\ \hline
\multirow{2}{*}{8}  & 16     & Teacher  & \multicolumn{3}{l}{Excellent. I believe you have a good understanding of Achilles now.}                                                                                                                                                                                                                  \\
                    & 17     & Student  & \multicolumn{3}{l}{Thanks, I think so too.}                                                                                                                                                                                                                                                              \\ \hline
9                   & 18     & Teacher  & \multicolumn{3}{l}{End of conversation.}                                                                                                                                                                                                                                                                 \\ \hline
\end{tabular}%
}
\end{table*}
\end{document}